\documentclass[journal]{IEEEtran}

\ifCLASSINFOpdf
\else
\fi

\RequirePackage{amsmath}
\usepackage{amsmath,amsfonts,amssymb}
\usepackage{graphicx}
\usepackage{siunitx}
\usepackage{cite}
\DeclareMathOperator*{\argmax}{argmax}
\usepackage{boldline,multirow}
\usepackage{color}
\usepackage{soul} 
\usepackage[table]{xcolor}
\usepackage[colorlinks=true,citecolor=blue,urlcolor=black]{hyperref}

\usepackage{xr}

\makeatletter
\newcommand*{\addFileDependency}[1]{
  \typeout{(#1)}
  \@addtofilelist{#1}
  \IfFileExists{#1}{}{\typeout{No file #1.}}
}
\makeatother

\newcommand*{\myexternaldocument}[1]{%
    \externaldocument{#1}%
    \addFileDependency{#1.tex}%
    \addFileDependency{#1.aux}%
}

\newcommand{\Reals}{\rm I\!R}

\myexternaldocument{supp}

\begin{document}
%


\title{Analyzing Overfitting under Class Imbalance in Neural Networks for Image Segmentation}

\author{Zeju~Li,
        Konstantinos~Kamnitsas
        and~Ben~Glocker
\thanks{Z. Li, K. Kamnitsas and B. Glocker are with the BioMedIA Group, Department of Computing, Imperial College London, SW72AZ, United Kingdom. E-mail: zeju.li18@imperial.ac.uk.}
}


\maketitle

\begin{abstract}

Class imbalance poses a challenge for developing unbiased, accurate predictive models. In particular, in image segmentation neural networks may overfit to the foreground samples from small structures, which are often heavily under-represented in the training set, leading to poor generalization. In this study, we provide new insights on the problem of overfitting under class imbalance by inspecting the network behavior. We find empirically that when training with limited data and strong class imbalance, at test time the distribution of logit activations may shift across the decision boundary, while samples of the well-represented class seem unaffected. This bias leads to a systematic under-segmentation of small structures. This phenomenon is consistently observed for different databases, tasks and network architectures. To tackle this problem, we introduce new asymmetric variants of popular loss functions and regularization techniques including a large margin loss, focal loss, adversarial training, mixup and data augmentation, which are explicitly designed to counter logit shift of the under-represented classes. Extensive experiments are conducted on several challenging segmentation tasks. Our results demonstrate that the proposed modifications to the objective function can lead to significantly improved segmentation accuracy compared to baselines and alternative approaches.

\end{abstract}


%

\IEEEpeerreviewmaketitle

\section{Introduction}
\label{sec1}

\IEEEPARstart{T}{he} success of convolutional neural networks (CNNs) is strongly linked with the availability of large scale, representative datasets. However, in many real-world applications such as medical image segmentation, the availability of large, annotated datasets is still limited. But even when there is a sufficient number of images available, the fundamental problem of class imbalance remains where region-of-interests (ROIs) (i.e. foreground classes) are heavily under-represented in the training data~\cite{brosch2016deep,litjens2017survey}. Similar to \cite{cui2019class}, the class imbalance ratio of one image can be defined as the ratio between the number of pixels of the background class (which is commonly the most frequent class) and the number of pixels of different object classes. The class imbalance ratio of a whole dataset would then be reported as the average class imbalance ratio of all images in the set. Class imbalance ratios of 100:1 or higher are not uncommon in applications such as lesion segmentation, as shown in Table~\ref{tab0}.

When the model is trained with imbalanced datasets, it can overfit to the training samples from the under-represented classes and may not generalize well during test time. However, the effects of overfitting under class imbalance on the model behavior is not well understood. In this study, we investigate how the distribution of activations of the classification layer (\emph{logits}) changes when the model is trained using different amounts of training data with strong class imbalance. As the model is trained with fewer training data and overfit the under-represented classes more, we find that the model projects unseen samples of the under-represented classes closer to and even across the decision boundary, while samples of the over-represented classes remain unaffected. This biased distribution shift leads to under-segmentation of under-represented class. Current solutions to address class imbalance or to mitigate overfitting do not explicitly consider this asymmetric logit shift and are unable to lead to significant improvements, as we show through an extensive set of experiments.

This study sheds new light on the problem of overfitting in the presence of class imbalance by making the following key contributions: 1) Via inspection of the network behavior on four segmentation tasks and datasets, and two popular model architectures, we conclude that overfitting under class imbalance consistently leads to decreased performance on under-represented classes specifically in terms of low sensitivity; 2) We identify the shift in the logit distribution of unseen test samples of under-represented classes as a result of overfitting under class imbalance; 3) Base on our observations, we propose simple yet effective asymmetric variants of five loss functions and regularization techniques which are explicitly designed to change the network behavior yielding improved segmentation accuracy for the under-represented classes.

This article is an extension of our earlier work presented at MICCAI 2019~\cite{li2019overfitting}. We extend our previous work on multiple aspects: 1) We provide a more detailed analysis including experiments on two additional datasets; 2) We further include a 3D U-Net to confirm that our observations hold across different network architectures; 3) We explore the proposed training objectives with the Dice loss in addition to cross-entropy; 4) We enrich the experiments by adding comparisons when training with F-score, extend experiments to multi-class segmentation, and also evaluate another regularization method which is noted as asymmetric augmentation. Our findings here confirm our initial observations about the biased behavior of neural networks. The behavior of logit distribution shift is consistently observed across different types of data, tasks and architectures. Our work highlights the importance of the issue of overfitting under class imbalance. The quantitative evaluation further supports our proposal of taking class imbalance into account when designing the learning objective.

\section{Related work}

\subsection{Class imbalance}

Class imbalance, which has been the focus of previous works~\cite{johnson2019survey, buda2018systematic}, is a common issue in image classification and image segmentation. Compared with the literature on class imbalance, a key contribution of this study is the focus on the model behaviour when it overfits to the under-represented classes with a detailed analysis and potential solutions. In the following, we discuss related work categorized by different methodological approaches.

\subsubsection{Re-weighting}

A common approach to tackle class imbalance is class-level re-weighting, which assigns higher weights or higher sampling probability to the under-represented classes based on sample frequency~\cite{zadrozny2003cost,wang2016training} or advanced rules~\cite{cui2019class}. In this study, explore re-weighting as a baseline approach in all experiments where we train the models with patches which are separately sampled from different classes with the same probability. Beyond that, sample-level re-weighting strategies are also proposed to build a balanced model. For example, hard sample mining was proposed to avoid the dominant effect of majority classes~\cite{dong2018imbalanced}. Similarly, focal loss and its variants were proposed to weight difficult samples over easy samples~\cite{hashemi2018asymmetric,lin2017focal,abraham2019novel, wong20183d} to steer the learning towards small objects. However, the under-represented samples are not necessarily difficult to predict during training. In fact, as we show empirically, the training samples of the under-represented classes are learned well due to overfitting. In this case, we find that a focal loss may even decrease the performance when processing imbalanced datasets because it reduces the focus on the under-represented samples. Therefore, in this study, we improve upon focal loss by removing the attenuation of under-represented classes. Margin based loss functions were proposed to learn discriminative embeddings and widely adopted to metric-learning and face recognition ~\cite{liu2016large, deng2019arcface}. Margin losses can also been seen as a kind of re-weighting approach which changes the magnitude of the gradient of the network output by multiplying a scalar, as we show in the supplementary Section~\ref{sec:marginlossanalysis}. In this study, we propose to only assign margins for the under-represented classes. The design of uneven margins for imbalanced datasets was first proposed in~\cite{li2002perceptron} for perception. Recently, Large Margin Local Embedding (LMLE) was proposed to put more constraints for the under-represented classes by only applying multiple margins to the minority classes, with a computationally expensive metric-learning based framework~\cite{huang2016learning}. More recently, two concurrent studies were also proposed to set larger margins for the under-represented classes from the perspective of uncertainty~\cite{khan2019striking} or generalization bound~\cite{zhou2018analysiscnns,cao2019learning}. In this study, we empirically show that one should not assign margins for the over-represented classes based on the observations of asymmetric logit distribution under class imbalance.

\subsubsection{Data synthesis}

Our work is related to data synthesis methods~\cite{chawla2002smote, guo2004learning} which generate synthetic samples of the minority classes based on intra-class relationship between samples to increase the variance of under-represented class. In addition, we create synthetic samples in the latent feature space rather than image space and provide two new ways to synthesize samples of the under-represented classes for modern machine learning models. We also propose to adopt stronger data augmentation for the under-represented classes by changing the augmentation probabilities to alleviate overfitting.

\subsubsection{Other methods}

The above mentioned methods are all based on changing the training data distribution to tackle class imbalance. In contrast, some other approaches try to counter class imbalance by modifying the training strategy. Specifically, ~\cite{havaei2017brain} firstly trained their model with data which is sampled from each class with the same probability. Thereafter, they only retrain the output layer with uniformly sampled data while freezing all other network parameters. In this way, they could separately learn a diverse representation and a classifier for realistic data distribution. Similar strategies, which aim to change the decision boundary at test-time, are also proposed recently for long-tailed recognition~\cite{zhou2020bbn,kang2019decoupling, tang2020long}. These approaches are complementary to our proposed solutions and could be combined. Other learning paradigms such as meta-learning~\cite{wang2017learning} and transfer learning~\cite{liu2019large} were also recently proposed for long-tail learning, but these are outside the scope of this paper.

\subsubsection{Segmentation}

The problem of class imbalance in image segmentation is different from that in image recognition~\cite{liu2019large} because the dominating class in image segmentation is the background class with diverse characteristics, and its segmentation accuracy is highly robust. In contrast, the accuracy for the majority classes in long-tailed image recognition can degrade with common techniques such as re-weighting~\cite{kang2019decoupling}. In addition, the evaluation of segmentation performance mostly relies on the foreground classes, and therefore, the focus is on improving accuracy in those classes. For example, recent studies proposed to provide a better trade-off between sensitivity and precision for segmentation~\cite{milletari2016v,hashemi2018asymmetric}. However, these strategies yield little improvements when processing highly imbalanced datasets, as shown in our experiments. This is because a deep neural network may achieve near-perfect training accuracy even for the under-represented samples without benefitting from the modified loss function. Class imbalance in image segmentation has been also approached via a boundary loss~\cite{kervadec2019boundary}. However, it is only applicable to segmentation. The authors show promising results with sufficient training data, but the model may still be prone to overfit the under-represented class with limited dataset. Other work adopted multi-stage approaches with candidate proposals and background suppression \cite{valindria2018small,setio2016pulmonary}. However, the candidate prediction process may still suffer from class imbalance. In addition, any missed candidates in one stage cannot be recovered in a later stage. In contrast, the solutions proposed here are general loss functions that can be incorporated into any model or learning approach and is applicable beyond image segmentation.

\subsection{Regularization techniques}

To improve generalization of deep neural networks, a number of regularization techniques are available. This includes dropout~\cite{srivastava2014dropout}, weight decay~\cite{krogh1992simple}, data augmentation~\cite{cubuk2019autoaugment,cubuk2019randaugment}, data mixing \cite{zhang2017mixup, yun2019cutmix}, and adversarial training~\cite{goodfellow2014explaining, xie2019adversarial}. However, most of these techniques were proposed for general image classification tasks where class imbalance is not explicitly addressed. It is also unclear how these techniques affect the network behavior in this setting.

\section{Overfitting under class imbalance and its effect on segmentation performance}
\label{sec:motivation}

\begin{figure*}[ht]
\centering
\includegraphics[width=\textwidth]{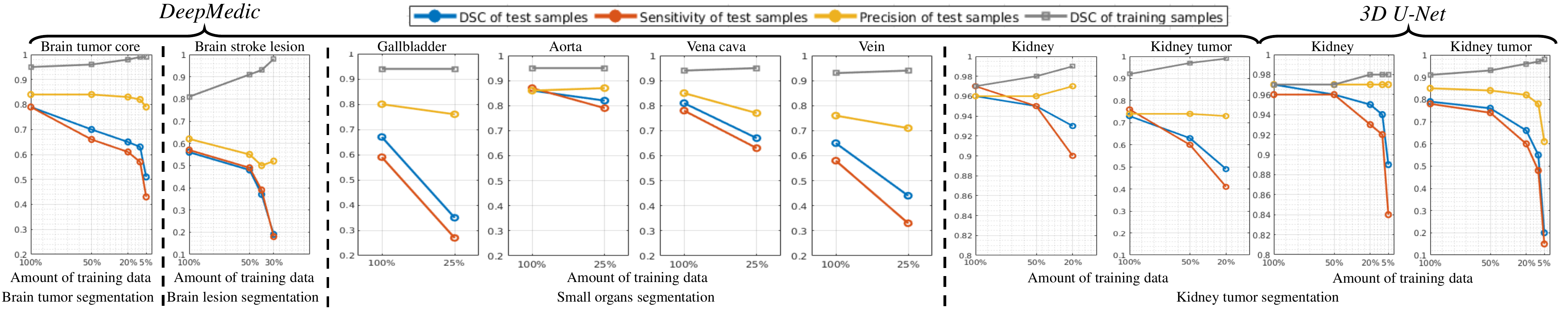}
\caption{Performance on brain tumor core, brain stroke lesion, small organs, kidney and kidney tumor segmentation with varying amounts of training data. The foreground (FG) and background (BG) samples are highly imbalanced, as noted below each subfigure. With less training data, performance drops due to the decrease of sensitivity, while the precision is largely retained.}
\label{fig1}
\end{figure*}

\begin{figure*}[ht]
\centering
\includegraphics[width=\textwidth]{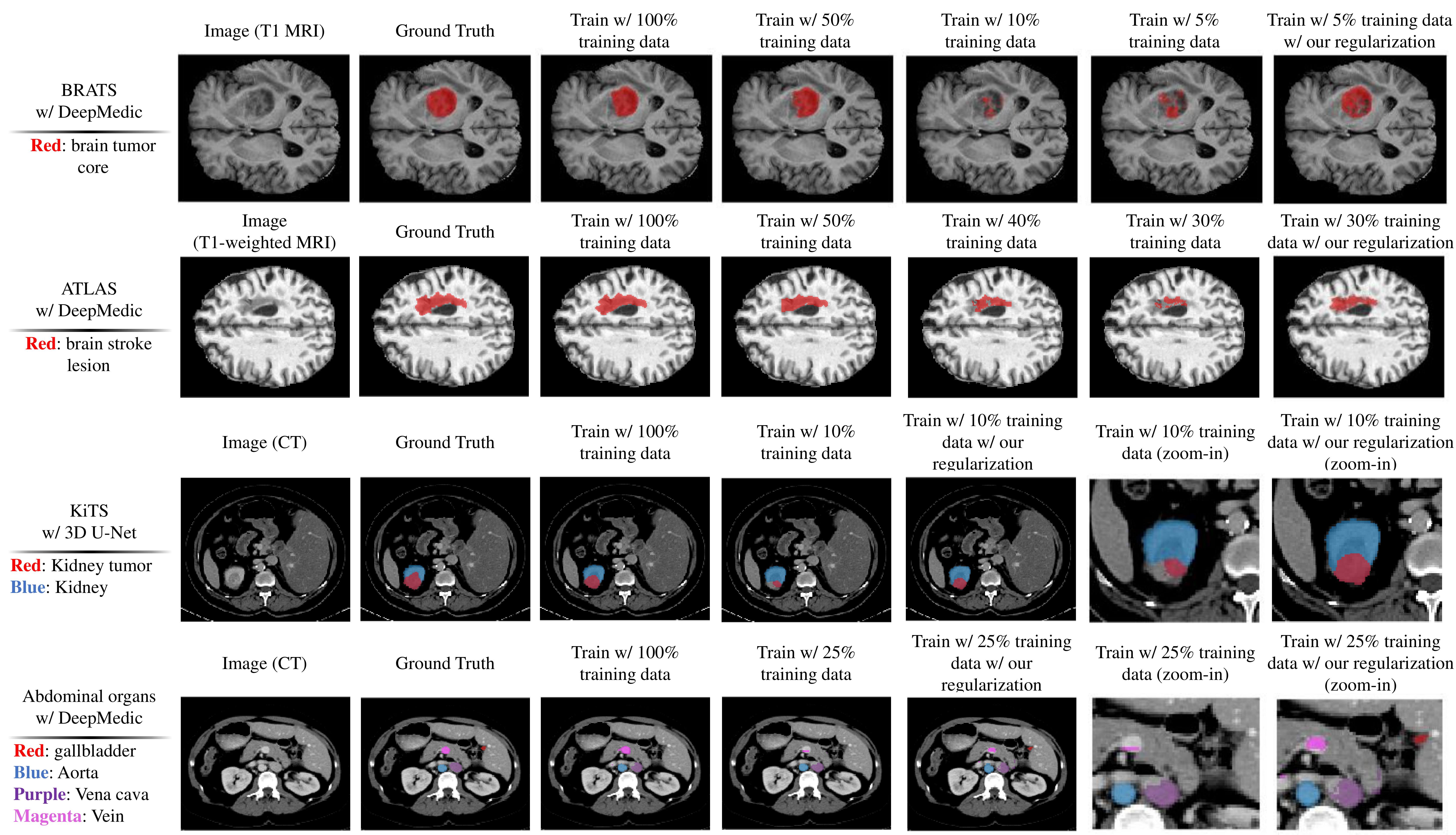}
\caption{Visualization of different datasets and segmentation results with different portions of training data. With less training data, the models are prone to under-segment the under-represented classes. The proposed regulation methods can alleviate the overfitting of under-represented classes and provide segmentation results with higher sensitivity and overall accuracy. Best viewed in color.}
\label{fig2visual}
\end{figure*}

To explore the effects of overfitting on the network behavior, we train CNNs using different amounts of data, on segmentation tasks that exhibit strong class imbalance. We conduct experiments on challenging segmentation tasks using data from the Multimodal Brain Tumor Image Segmentation (BRATS) challenge \cite{bakas2017advancing}, the Anatomical Tracings of Lesions After Stroke (ATLAS) dataset~\cite{liew2018large}, small organ segmentation (data from~\cite{xu2015efficient}) and Kidney Tumor Segmentation (KiTS)~\cite{heller2019kits19}. The statistics of those four datasets are summarized in Table~\ref{tab0}. To ensure our findings generalize across models, in our investigation we employ two convolutional network architectures that have been proven potent in a variety of segmentation tasks: We employ a DeepMedic architecture~\cite{kamnitsas2017efficient} for the experiments on brain lesions and multi-organ segmentation tasks on which it has previously shown high performance~\cite{kamnitsas2017efficient,savenije2020clinical}, and a well configured 3D U-Net~\cite{isensee2019automated} for the experiments on kidney tumor segmentation on KiTS19 data, which is the base model of the winning entry of KiTS19 challenge~\cite{heller2019kits19}. The detailed network configurations are summarized in Section~\ref{sec4}.

\begin{figure*}[t]
\centering
\includegraphics[width=\textwidth]{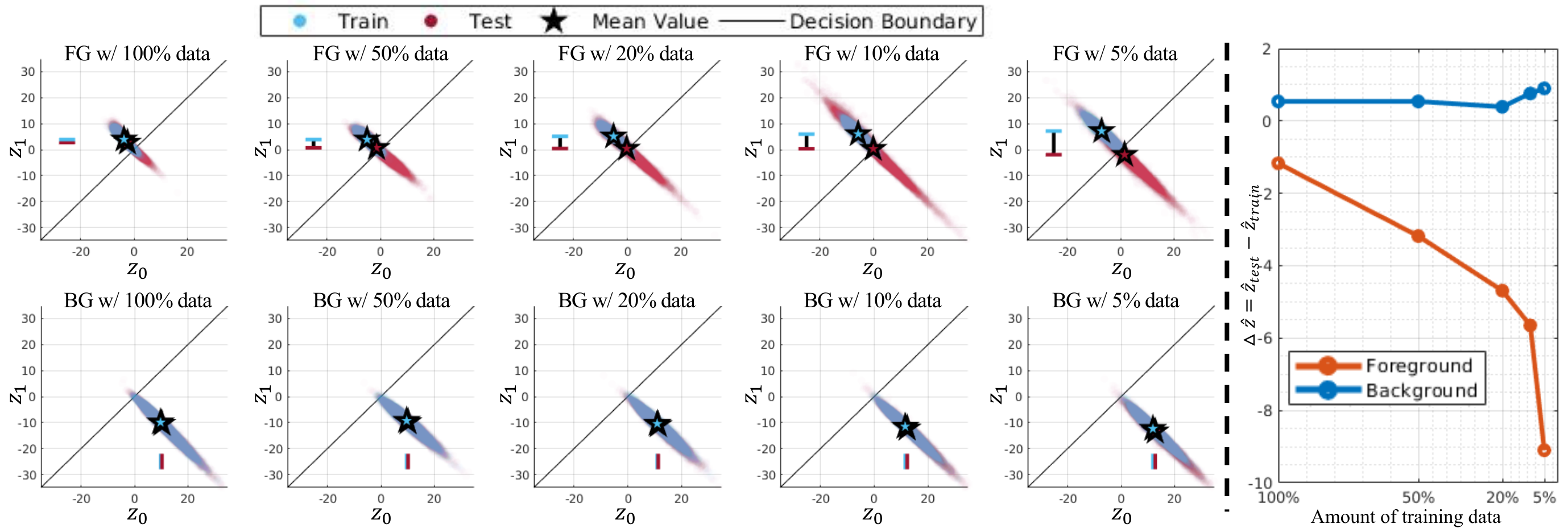}
\caption{(Left part) Activations of the classification layer (logit $z_{0}$ for background, logit $z_{1}$ for brain tumor core) when processing (top) tumor and (bottom) background samples of BRATS with DeepMedic, using different amounts of training data. The CNN maps training and testing samples of the background class to similar logit values. However, mean activation for testing data shifts significantly for the tumor class towards and sometimes across the decision boundary. (Right part) The shift of mean value of logits observed when processing training and testing data ($ \Delta \hat{z} = |\hat{z}_{test}| - |\hat{z}_{train}| $). }
\label{fig2}
\end{figure*}

\begin{figure*}[t]
\centering
\includegraphics[width=\textwidth]{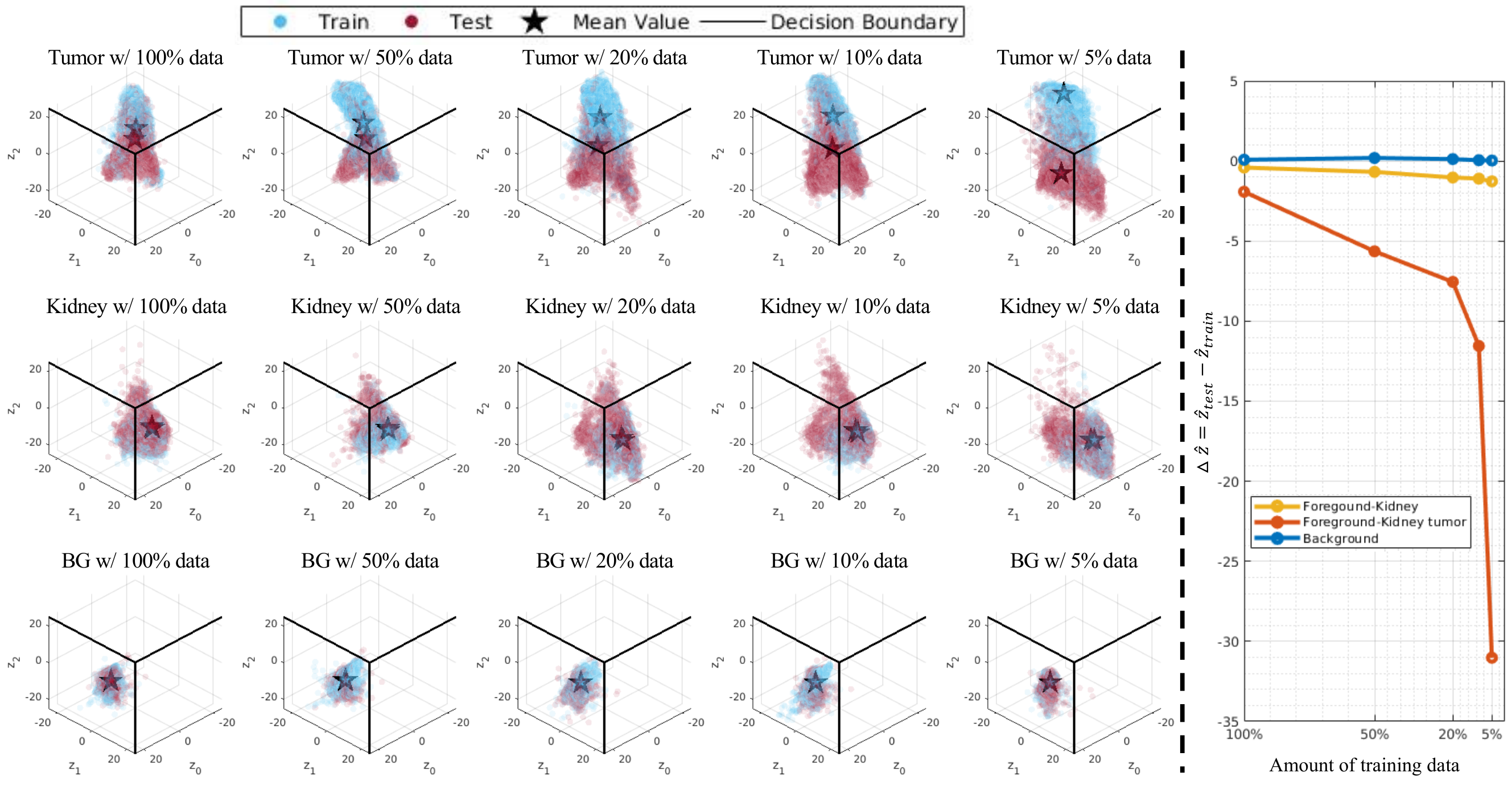}
\caption{(Left part) Activations of the classification layer (logit $z_{0}$ for background, logit $z_{1}$ for kidney, logit $z_{2}$ for kidney tumor) when processing (top) tumor, (middle) kidney and (bottom) background samples of KiTS with 3D U-Net, using different amounts of training data. The CNN also fails to map the training and testing samples of the tumor class in a similar position. (Right part) The shift of mean value of logits.}
\label{fig3}
\end{figure*}

\begin{figure}[t]
\centering
\includegraphics[width=0.48\textwidth]{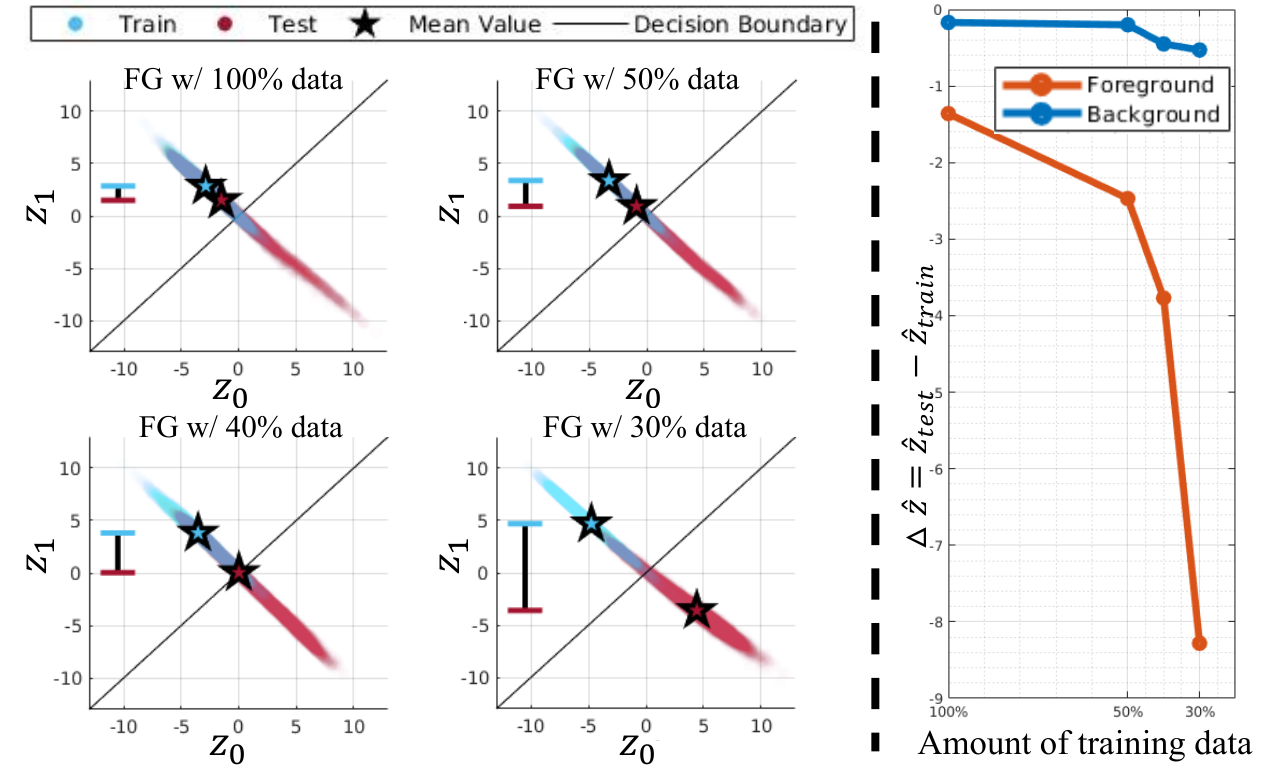}
\caption{(Left part) Activations of the classification layer when processing lesion samples of ATLAS with DeepMedic, using different amounts of training data. (Right part) The shift of mean value of logits.}
\label{fig4}
\end{figure}

\begin{table}[ht]
\centering
\caption{The statistics and class imbalance ratios of the four datasets used in this study. Class imbalance ratio is defined as the average ratio between the number of the background (BG) pixels and the foreground (FG) pixels over all images.}
\label{tab0}
\newsavebox{\tableboxh}
\begin{lrbox}{\tableboxh}
\begin{tabular}{c|c|c|c}
\hlineB{3}
Dataset & \multicolumn{1}{m{14mm}<{\centering}}{Total FG pixels} & \multicolumn{1}{|m{17mm}<{\centering}}{Total BG pixels} & \multicolumn{1}{|m{18mm}<{\centering}}{Class imbalance ratio (avg. $\pm$ std.)} \\
\hlineB{2}
BRATS & 1.2 $\times$ 10$^7$ & 253.2 $\times$ 10$^7$ & 712.8 $\pm$ 1463.1 \\
ATLAS & 4.6 $\times$ 10$^6$ & 1908.6 $\times$ 10$^6$ & 1768.7 $\pm$ 6710.8 \\
KiTS-Kidney & 13.5 $\times$ 10$^6$ & 1662.0 $\times$ 10$^6$ & 122.8 $\pm$ 69.3  \\
KiTS-Tumor & 2.9 $\times$ 10$^6$ & 1662.0 $\times$ 10$^6$ & 6736.6 $\pm$ 1522.9 \\
Abdomen-Gallbladder & 1.0 $\times$ 10$^5$ & 2630.5 $\times$ 10$^5$ & 4887.5 $\pm$ 4854.7 \\
Abdomen-Aorta & 3.8 $\times$ 10$^5$ & 2630.5 $\times$ 10$^5$ & 829.2 $\pm$ 686.1 \\
Abdomen-Vena cava & 3.4 $\times$ 10$^5$ & 2630.5 $\times$ 10$^5$ & 832.2 $\pm$ 278.2  \\
Abdomen-Vein & 1.4 $\times$ 10$^5$ & 2630.5 $\times$ 10$^5$ & 2196.5 $\pm$ 792.2 \\
\hline
\end{tabular}
\end{lrbox}
\scalebox{0.95}{\usebox{\tableboxh}}
\end{table}

The observations on the test and training set are summarized in Fig.~\ref{fig1}. With less training data, we notice a clear decrease of segmentation accuracy on test data while the accuracy on training data increases due to easier overfitting, as expressed by DSC (defined as $ DSC=2\frac{\text{sensitivity} \cdot \text{precision}}{\text{sensitivity} + \text{precision}}$). We observe that overfitting leads to a reduction of sensitivity while precision remains largely stable. In all settings and tasks, the specificity of the foreground always remains near-perfect (\textgreater0.999) with different amounts of training data, indicating the predictions of background samples are stable (not shown in Fig.~\ref{fig1} to avoid cluttering). We also provide some corresponding visualization examples in Fig~\ref{fig2visual}.

Our observations on four different datasets show that this behavior is consistent when foreground classes are under-represented and is not specific to particular tasks and model architectures. Our findings reveal that models that overfit to imbalanced training data have a bias to under-segment the under-represented class on unseen test data.

\subsection{Logit distribution shift}

To obtain a better understanding of the network behavior after training on imbalanced data, we monitor the logit distribution when processing training and unseen test samples. The observations we make for the tasks of brain tumor core, kidney tumor and brain stroke lesion segmentation are summarized in Fig.~\ref{fig2}, \ref{fig3} and \ref{fig4}, respectively. We notice that the logit distribution of foreground samples shifts significantly towards and even across the decision boundary, while the logit distribution of background samples remains stable. The shift of the foreground logits results in a higher number of false negatives, which causes a drastic decrease of sensitivity (calculated as $\frac{\text{TP}}{\text{TP} + \text{FN}}$). This biased logit shift under class imbalance may also occur in other tasks such as image classification. However, it is particularly prevalent in image segmentation with small structures-of-interest.

We find that this shift of logits correlates with how much a model overfits to the under-represented class. Training with less data leads to more overfitting, and the logit distribution shift becomes larger. Moreover, we find that the logit shift also correlates with the size of structures represented by the foreground class. The rarest class shifts the most, as shown in the right part of Fig.~\ref{fig3}.

In image segmentation, a CNN is optimized to push the logits of different classes away from each other and far from the decision boundary. It is relatively easy for a deep CNN to build an embedding for the training samples from the under-represented class because it just needs to build a set of case-specific filters to facilitate memorization. For example, as a CNN will only observe very few training samples of the foreground class, a CNN can dedicate specific model parameters to memorize all foreground samples, even if the individual patterns are rather complex. Specifically, we find a CNN seems to be more confident about foreground samples during training, mapping them farther away from the decision boundary when overfitting, as shown in Fig.~\ref{fig2}, Fig.~\ref{fig3} and Fig.~\ref{fig4}. However, these tailored filters will not generalize to unseen test data. Therefore, the activations for test samples of the under-represented class are smaller in magnitude (sub-optimal pattern matching of filters and unseen samples), leading to the observed distribution shift. In contrast, a CNN has to build generic filters for a well represented class to represent many different characteristics of the same class, leading to good generalization. Such filters will map unseen samples to similar locations in logit space and no shift between the embeddings of training and test samples is observed. As a result of class imbalance \emph{and} overfitting, a CNN may underperform on the under-represented class while still generalizing well for the well-represented class.

While the negative effect of class imbalance and overfitting on model performance is well known, to our knowledge there has been little work investigating the specifics how the network behavior is affected. Only by understanding better the implication, we can devise mitigation strategies. Previous loss functions and regularization techniques that aim to prevent overfitting did not take the behavior that we observe into account, and thus show limited success for improving segmentation accuracy in the setting of limited data with strong class imbalance. Here, we propose solutions via new asymmetric variants of existing objective functions leading to better feature embeddings for the under-represented samples, leading to significant improvements in segmentation accuracy for small structures-of-interest.

\begin{figure*}[t]
\centering
\includegraphics[width=\textwidth]{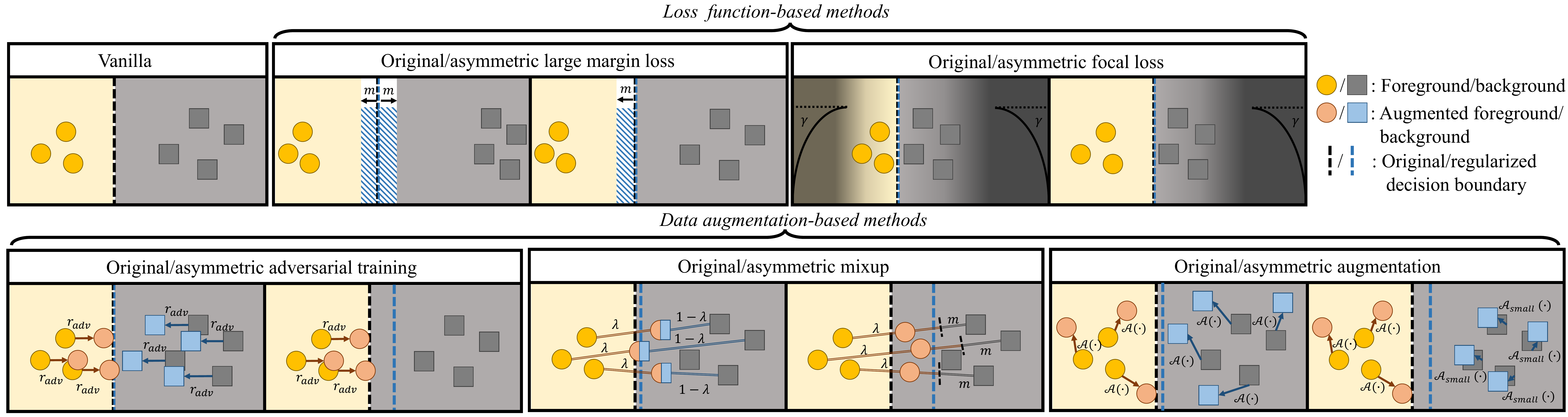}
\caption{The illustration of the proposed asymmetric modifications for the existing loss functions and regularization techniques. We make the logit activations of foreground class far away from the decision boundary by setting a bias for the foreground class in different ways.} \label{fig5}
\end{figure*}

\section{Tackling overfitting under class imbalance with asymmetric objective functions}
\label{sec3}
Based on our observations above about the biased behavior of CNNs, we design modifications to existing loss functions and training strategies to prevent the logit distribution shift. Specifically, we add a bias for the under-represented class. Although the original techniques were proposed for different purposes, our modifications share a common goal: keep the logit activations of the under-represented class away from the decision boundary. Even if the logit of a foreground sample shifts towards the decision boundary as long as it does not cross it, its prediction remains correct (cf. Fig.~\ref{fig5}).

\subsection{Asymmetric large margin loss}
\label{sec:margin}
We consider a CNN for the task of semantic segmentation. For a training dataset $\{ (\boldsymbol{x}_i, \boldsymbol{y}_i) \}_{i = 1}^N$ with ${N}$ samples, we denote a training sample with $\boldsymbol{x}_i$ and its corresponding one-hot vector $\boldsymbol{y}_{i}$. If $c$ is the total number of classes of the task, $\boldsymbol{y}_{i}$ has $c$ elements, with its $j$'th element $y_{ij} \in \{0,1\}$ corresponding to the $j$'th class. $y_{ij}$ equals to $1$ if $j$ is the real class of $\boldsymbol{x}_i$, or $0$ otherwise. With this notation,  the cross-entropy (CE) loss can be written as\footnote{We formulate CE as sum over classes to make class specific modifications.}:

\begin{equation}
    L_{CE}(\boldsymbol{x_{i}},\boldsymbol{y_{i}}) = -\sum_{j=1}^{c} y_{ij}\log(p_{ij}),
    \label{eq:CE}
\end{equation}

where $p_{ij}$ is the predicted probability by the network that the real class of $\boldsymbol{x}_i$ is $j$. Probability $p_{ij}$ is commonly obtained via a softmax function over the $c$ activations $\{(z_{ij})_{j=1}^{c} \in 
\Reals^c\}$ that the network outputs for $\boldsymbol{x}_i$ at its last layer. These activations are called the \emph{logits}. With this, $p_{ij}$ is given by:

\begin{equation}
    p_{ij} = \frac{\mathrm{e}^{z_{ij}}}{\sum_{j=1}^{c}\mathrm{e}^{z_{ij}}}.
    \label{eq:prob}
\end{equation}

Besides CE, the smooth version of the DSC metric is an alternative choice for the loss function which is widely used for medical image segmentation~\cite{milletari2016v}. DSC loss can be calculated in the form of $1 - \text{DSC}=\frac{\text{FP} + \text{FN}}{2\,\text{TP} + \text{FP} + \text{FN}}$), which is:

\begin{equation}
\resizebox{1\hsize}{!}{$
\begin{aligned}
    L_{DSC}(\boldsymbol{x_{i}},\boldsymbol{y_{i}}) =\sum_{j=1}^{c}\Big(\frac{(1-y_{ij})p_{ij}+y_{ij}(1-p_{ij})}{2y_{ij}p_{ij}+(1-y_{ij})p_{ij}+y_{ij}(1-p_{ij})}\Big).
    \label{eq:DSC}
\end{aligned}$}
\end{equation}

The large margin loss was proposed for increasing the Euclidean distances between logits for different classes to learn discriminative features \cite{wang2018additive}. Symmetrically, it is implemented by adding a margin on the logits of every class:

\begin{equation}
    L_{CE_M}(\boldsymbol{x_{i}},\boldsymbol{y_{i}}) = -\sum_{j=1}^{c} y_{ij}\log(q_{ij}),
    \label{eq:margin}
\end{equation}

in which we require: 

\begin{equation}
    q_{ij} = \frac{\mathrm{e}^{z_{ij} - y_{ij}m}}{\sum_{j=1}^{c}\mathrm{e}^{z_{ij} - y_{ij}m}},
    \label{eq:marginprob}
\end{equation}

where $m$ is a hyper-parameter for the margin. Although the large margin loss encourages the model to map different classes away from each other, the decision boundary remains in the center. According to our observations, class imbalance causes shifts of unseen foreground samples towards the background class. To mitigate this, a regularizer may aim to move the decision boundary closer to the background class. Our asymmetric modification only sets the margin for the rare classes. We define $\boldsymbol{r}$ as a one-hot vector with $c$ elements, with its $j$'th element $r_{j} \in \{0,1\}$ corresponding to the $j$'th class and $r_j$ equals to 1 if $j$ is taken as the rare class. With the indication of $\boldsymbol{r}$, we derive the asymmetric large margin loss as:

\begin{equation}
    \hat{L}_{CE_M}(\boldsymbol{x_{i}},\boldsymbol{y_{i}}) = -\sum_{j=1}^{c} y_{ij}\log(\hat{q}_{ij}),
    \label{eq:modified margin}
\end{equation}

where we require:

\begin{equation}
    \hat{q}_{ij} = \frac{\mathrm{e}^{z_{ij} - y_{ij}r_jm}}{\sum_{j=1}^{c}\mathrm{e}^{z_{ij} - y_{ij}r_jm}},
    \label{eq:marginprobasy}
\end{equation}

In this study, we define $r_j$ as 1 for the foreground samples and 0 for the background samples. In other applications, $r_j$ can also be defined as a continuous variable indicating the rarity of the classes with $r_j\in[0,1]$, for methods in Section~\ref{sec:margin}, \ref{sec:focal} and \ref{sec:advsarial}. Similarly, the symmetric and asymmetric large margin loss for DSC loss can be derived by substituting equation~\ref{eq:marginprob}:

\begin{equation}
\resizebox{1\hsize}{!}{$
\begin{aligned}
    L_{DSC_M}(\boldsymbol{x_{i}},\boldsymbol{y_{i}}) =\sum_{j=1}^{c}\Big(\frac{(1-y_{ij})q_{ij}+y_{ij}(1-q_{ij})}{2y_{ij}q_{ij}+(1-y_{ij})q_{ij}+y_{ij}(1-q_{ij})}\Big),
    \label{eq:DSCmargin}
\end{aligned}$}
\end{equation}

and

\begin{equation}
\resizebox{1\hsize}{!}{$
\begin{aligned}
    & \hat{L}_{DSC_M}(\boldsymbol{x_{i}},\boldsymbol{y_{i}}) =\sum_{j=1}^{c}\Big(\frac{(1-y_{ij})\hat{q}_{ij}+y_{ij}(1-\hat{q}_{ij})}{2y_{ij}\hat{q}_{ij}+(1-y_{ij})\hat{q}_{ij}+y_{ij}(1-\hat{q}_{ij})}\Big).
    \label{eq:DSCmodifiedmargin}
\end{aligned}$}
\end{equation}

\subsection{Asymmetric focal loss}
\label{sec:focal}
The focal loss was proposed for small object detection by reducing the weight for well-classified samples and focusing on samples which are near the decision boundary~\cite{lin2017focal}. It adds attenuation inside the loss function based on the logit activations:

\begin{equation}
    L_{CE_{focal}}(\boldsymbol{x_{i}},\boldsymbol{y_{i}}) = -\sum_{j=1}^{c} (1-p_{ij})^\gamma y_{ij}\log(p_{ij}),
    \label{eq:Focal}
\end{equation}

where $\gamma$ is the hyper-parameter to control the focus. The symmetric focal loss prevents logits from being too large and makes every class stay near the decision boundary. However, this makes it likely for the unseen foreground samples to shift across the decision boundary. We remove the loss attenuation for the foreground class to keep it away from the decision boundary:

\begin{equation}
\begin{split}
    & \hat{L}_{CE_{focal}}(\boldsymbol{x_{i}},\boldsymbol{y_{i}}) = \sum_{j=1}^{c}\Big(-r_j y_{ij}\log(p_{ij}) \\
    & - (1-r_j)(1-p_{ij})^\gamma y_{ij}\log(p_{ij})\Big).
    \label{eq:modified focal}
\end{split}
\end{equation}

Inspired by the focal loss~\cite{lin2017focal}, related work integrates a similar attenuation term into the DSC loss~\cite{abraham2019novel, wong20183d}. In practice, we find that the logarithmic DSC loss~\cite{abraham2019novel} significantly changes the magnitude of DSC loss making it difficult to be combined with other losses. The attenuation in the focal Tversky loss~\cite{wong20183d} is very large and may overly suppress the easier class. Here, we propose another form of DSC loss with an adaptive weight preserving a similar loss magnitude. Specifically, we add the attenuation term to the false negatives part of the function and prevent the network being too confident about its prediction:

\begin{equation}
\resizebox{1\hsize}{!}{$
\begin{aligned}
    L_{DSC_{focal}}(\boldsymbol{x_{i}},\boldsymbol{y_{i}}) =\sum_{j=1}^{c}\Big(\frac{(1-y_{ij})p_{ij}+(1-p_{ij})^\gamma y_{ij}(1-p_{ij})}{2y_{ij}p_{ij}+(1-y_{ij})p_{ij}+y_{ij}(1-p_{ij})}\Big),
    \label{eq:DSCfocal}
\end{aligned}$}
\end{equation}

Compared with the original version of the CE loss, this formulation for the DSC loss has a similar effect of reducing the penalty for the well-classified samples while keeping the magnitude of the loss similar to the original one, as shown in Supplementary Fig.~\ref{figa6}. We refer to this as the focal DSC loss in the following. Similarly, the asymmetric version of the focal DSC loss is derived by removing the attenuation term for the foreground class:

\begin{equation}
\resizebox{1\hsize}{!}{$
\begin{aligned}
    & \hat{L}_{DSC_{focal}}(\boldsymbol{x_{i}},\boldsymbol{y_{i}}) =\sum_{j=1}^{c}\Big(\frac{(1-y_{ij})p_{ij}+r_jy_{ij}(1-p_{ij})}{2y_{ij}p_{ij}+(1-y_{ij})p_{ij}+y_{ij}(1-p_{ij})}\\
    &+\frac{(1-y_{ij})p_{ij}+(1-r_j)(1-p_{ij})^\gamma y_{ij}(1-p_{ij})}{2y_{ij}p_{ij}+(1-y_{ij})p_{ij}+y_{ij}(1-p_{ij})}\Big).
    \label{eq:DSCmodifiedfocal}
\end{aligned}$}
\end{equation}

\subsection{Asymmetric adversarial training}
\label{sec:advsarial}

Adversarial training was proposed to learn more robust classifiers by training with difficult samples~\cite{goodfellow2014explaining}. The network is trained by considering adversarial samples as additional training data~\cite{xie2019adversarial, miyato2018virtual}:

\begin{equation}
    L_{adv}(\boldsymbol{x_{i}},\boldsymbol{y_{i}}) = L(\boldsymbol{x_{i}},\boldsymbol{y_{i}}) + L(\boldsymbol{x_{i}}+l \cdot \frac{\boldsymbol{d_{adv}}}{\|\boldsymbol{d_{adv}}\|_2},\boldsymbol{y_{i}}),
    \label{eq:Adversarial training}
\end{equation}

\begin{equation}
    \textrm{with}\quad  \boldsymbol{d_{adv}} = \argmax_{\boldsymbol{d};\|\boldsymbol{d}\|<\epsilon}L(\boldsymbol{x_{i}}+\boldsymbol{d},\boldsymbol{y_{i}}).
    \label{eq:Adversarial direction}
\end{equation}

Here, $\boldsymbol{d_{adv}}$ is the direction of the generated adversarial samples, $l$ and $\epsilon$ are the magnitude and the range of the adversarial perturbations, respectively. $L$ is the chosen loss function, which can be $L_{CE}$ and / or $L_{DSC}$. Similar to the large margin loss, symmetric adversarial training preserves the decision boundary and may cause difficulties for unseen foreground samples, which tends to shift towards background class. Our proposed asymmetric adversarial training aims to produce a larger space between the foreground class and the decision boundary. Specifically, we generate samples by considering more from the rare classes:

\begin{equation}
    \boldsymbol{\hat{d}_{adv}} = \argmax_{\boldsymbol{d};\|\boldsymbol{d}\|<\epsilon}L(\boldsymbol{x_{i}}+\boldsymbol{d},\boldsymbol{y_{i}} \odot \boldsymbol{r} )\Big|_{\boldsymbol{y_{i}} \cdot \boldsymbol{r}>0},
    \label{eq:Adversarial direction2}
\end{equation}

where ``$\odot$'' refers to the element product and ``$\cdot$'' refers to the dot product.

\subsection{Asymmetric mixup}
Mixup is a simple yet effective data augmentation algorithm to improve generalization by generating extra training samples by using the linear combination of pairs of images and their labels \cite{zhang2017mixup}:

\begin{equation}
    L_{mixup}(\boldsymbol{x_{i}},\boldsymbol{y_{i}},\boldsymbol{x_{k}},\boldsymbol{y_{k}}) = L(\boldsymbol{x_{i}},\boldsymbol{y_{i}}) + L(\boldsymbol{\tilde{x_{i}}},\boldsymbol{\tilde{y_{i}}}),
    \label{eq:mixup}
\end{equation}

where ($\boldsymbol{\tilde{x_{i}}}$, $\boldsymbol{\tilde{y_{i}}}$) is the generated training sample:

\begin{equation}
    \boldsymbol{\tilde{x_{i}}} = \lambda \boldsymbol{x_{i}}+(1-\lambda)\boldsymbol{x_{k}}, \quad \boldsymbol{\tilde{y_{i}}} = \lambda \boldsymbol{y_{i}}+(1-\lambda)\boldsymbol{y_{k}}.
    \label{eq:mix_x}
\end{equation}

Here, $\lambda$ is randomly selected based on a beta distribution, ($\boldsymbol{x_{k}}$, $\boldsymbol{y_{k}}$) is another random training sample. Mixup regularizes the model by centering the decision boundary between classes which helps very little in our setting. Different from the original mixup, which generates samples with soft labels, our modification generates hard labels by considered augmented samples near to the foreground samples as foreground class. Asymmetric mixup can keep the decision boundary away from the foreground class and increase the area of the foreground logit distribution. This prevents unseen under-presented samples from shifting across the decision boundary. Specifically, the mixed image $\tilde{\boldsymbol{x_{i}}}$ which has a certain distance from the background class, is taken as a foreground sample:

\begin{equation}
\resizebox{1\hsize}{!}{$
\begin{aligned}
\boldsymbol{\hat{\tilde{y_{i}}}}=\left\{
\begin{array}{rcl}
\boldsymbol{y_{i}} &      & {\text{if}\ \big(\lambda > m \quad \textrm{and} \quad \boldsymbol{y_{i}} \cdot \boldsymbol{r}(1-\boldsymbol{y_{k}} \cdot \boldsymbol{r}) == 1\big)  \quad \textrm{or} \quad \boldsymbol{y_{i}} == \boldsymbol{y_{k}} },\\
\boldsymbol{y_{k}}       &      & {\text{if}\ \big(1-\lambda > m \quad \textrm{and} \quad \boldsymbol{y_{k}} \cdot \boldsymbol{r}(1-\boldsymbol{y_{i}} \cdot \boldsymbol{r}) == 1\big) \quad \textrm{or} \quad \boldsymbol{y_{i}} == \boldsymbol{y_{k}} },\\
\boldsymbol{0}       &      & \textrm{otherwise},
\end{array} \right.
\label{eq:mix_yRE}
\end{aligned}$}
\end{equation}

where $\emph{m}$ is the margin to guarantee that the augmented samples are not getting too close to background samples. In practice, we do not update the model using training samples with $\boldsymbol{\hat{\tilde{y_{i}}}} = \boldsymbol{0}$.

\subsection{Asymmetric augmentation}

In order to extend the latent space of the foreground class, we also evaluate a simple method to compensate class imbalance by adjusting the magnitude of augmentation for different classes. Standard data augmentation methods would preserve the label $\boldsymbol{y_i}$ and adopt the same set of heuristic transformations such as scaling and rotations to the original training sample $\boldsymbol{x_i}$ for different classes. The generated training sample $\boldsymbol{\tilde{x_{i}}}$ can be obtained using:

\begin{equation}
    \boldsymbol{\tilde{x_{i}}} = \mathcal{A}(\boldsymbol{x_{i}}),
    \label{eq:augment_x}
\end{equation}

where $\mathcal{A}$ is the chosen transformation with certain probability. When the dataset is highly imbalanced, adding more synthesized background samples is not necessary. Our simple variant of data augmentation reduces the number of transformed samples for the background classes. In this asymmetric setting, the generated sample $\tilde{x_{i}}$ is obtained using:

\begin{equation}
\boldsymbol{\hat{\tilde{x_{i}}}}=\left\{
\begin{array}{rcl}
\mathcal{A}(\boldsymbol{x_{i}})         &      & {\text{if}\ \quad \boldsymbol{y_{i}} \cdot \boldsymbol{r} == 1},\\
\mathcal{A}_{small}(\boldsymbol{x_{i}})       &      & \textrm{otherwise},
\end{array} \right.
\label{eq:augment_asy}
\end{equation}
where $\mathcal{A}_{small}$ is transformations with smaller probability.

\subsection{The combination of asymmetric techniques}

The above-mentioned modifications would introduce more variances for the under-represented classes in the latent space or the image space by adding a bias for the foreground class from different perspectives. In practice, some or all of the techniques can be integrated into a single model to combat overfitting under class imbalance.

Specifically, we can first generate different sets of the augmented samples following the asymmetric adversarial training, the asymmetric mixup and the asymmetric augmentation following equation~\ref{eq:Adversarial direction2}, \ref{eq:mix_yRE} and \ref{eq:augment_asy}, separately. The network can then be optimized using the extended training set with the loss functions combined with the asymmetric large margin loss and asymmetric focal loss:

\begin{equation}
\begin{split}
    & \hat{L}_{CE_{combine}}(\boldsymbol{x_{i}},\boldsymbol{y_{i}}) = \sum_{j=1}^{c} \Big(-r_j y_{ij}\log(\hat{q}_{ij})\\
    & -(1-r_j)(1-\hat{q}_{ij})^\gamma y_{ij}\log(\hat{q}_{ij})\Big).
    \label{eq:comb}
\end{split}
\end{equation}

A combined DSC loss can be formulated in a similar way.

\section{Experiments}
\label{sec4}

\subsection{Experimental setup}

We demonstrate the effect of our proposed modifications with a variety of medical image segmentation tasks using different models and training scenarios. Here, we summarize the dataset splits and experimental settings, which are kept the same with motivational experiments in Section~\ref{sec:motivation}. We keep the hyper-parameters of the methods the same for the original baselines and our modified techniques. The hyper-parameters are summarized in Supplementary Table~\ref{taba1}, \ref{taba2} and \ref{taba3}. Additionally, we conduct a sensitivity analysis of all the hyper-parameters and summarize the results in Supplementary Table~\ref{taba7}. We also provide the source code for our experiments\footnote{\url{https://github.com/ZerojumpLine/OverfittingUnderClassImbalance}}.

\subsubsection{Brain tumor segmentation}

We first evaluate the asymmetric techniques for the case of binary brain tumor core segmentation using the DeepMedic network architecture, a well performing method for this task \cite{kamnitsas2017efficient}. To investigate the behavior under overfitting and to isolate better the effect of the objective functions, we do not use dropout, weight decay and data augmentation in this experiment. We train the network with CE loss, unless otherwise specified. By default, we sample 50\% training samples from the foreground class. We conduct experiments using the training dataset of BRATS2017 dataset~\cite{bakas2017advancing} which contains 285 four modalities Magnetic Resonance (MR) images. The MR images all have the same voxel space of 1.0$\times$1.0$\times$1.0 mm. We test on 95 cases and train separate models using 190 (100\%), 95 (50\%), 38 (20\%), 19 (10\%) and 10 cases (5\% of full training set). 

\subsubsection{Brain stroke lesion segmentation}

We also evaluate the asymmetric techniques for the case of brain stroke lesion segmentation \cite{liew2018large} again using DeepMedic. Here, we use a more realistic setting, employing standard regularization techniques including dropout, weight decay and data augmentation, as in the original work where the model achieved high performance for stroke lesion segmentation \cite{kamnitsas2017efficient}. We implement our asymmetric techniques with the default training setting and default network architecture. The augmentation includes small intensity shifts and flipping in the sagittal plane wFith probability 0.5. The network is always trained with CE loss. We conduct experiments using ATLAS dataset~\cite{liew2018large} which contains 220 T1-weighted MR images. The MR images have the same voxel space of 1.0$\times$1.0$\times$1.0 mm. We test on 75 cases and train separate models using 145 (100\%), 73 (50\%), 57 (40\%) and 43 cases (30\% of full training set). 

\subsubsection{Small organ segmentation}
For organ segmentation in Section~\ref{sec:motivation}, we use a default DeepMedic network. We conduct experiments using the training datatset of the abdominal organ segmentation challenge~\cite{xu2015efficient} which contains 30 computed tomography (CT) scans. We train the network to segment thirteen abdominal organs. We test on 10 cases and train models using 20 (100\%) and 5 cases (25\% of training set). We resample all the MR images to a common voxel spacing of 2.0$\times$2.0$\times$2.0 mm. We show the segmentation results of representative small organs including gallbladder, aorta, inferior vena cava as well as portal vein and splenic vein in Fig.~\ref{fig1}. Here, we use this dataset to further confirm the observations about the effect on sensitivity when training with varying amounts data.

\begin{table*}[t]
\centering
\caption{Evaluation of brain tumor core segmentation using DeepMedic with different amounts of training data and different techniques to counter overfitting. The results are calculated with post-processing. Results which have worse DSC than the vanilla baseline are highlighted with shading. Best and second best results are in bold with the best also underlined.}\label{tab1}
\newsavebox{\tablebox}
\begin{lrbox}{\tablebox}
\begin{tabular}{p{38mm}<{\centering}|c|c|c|c||c|c|c|c||c|c|c|c||c|c|c|c}
\hlineB{3}
\multirow{2}{*}{Method} & \multicolumn{4}{c||}{5\% training} & \multicolumn{4}{c||}{10\% training} & \multicolumn{4}{c||}{20\% training} & \multicolumn{4}{c}{50\% training}  \\ 
 &  DSC & SEN &  PRC & HD &   DSC & SEN &  PRC & HD &   DSC & SEN &  PRC & HD &   DSC & SEN &  PRC & HD  \\
\hlineB{2}
Vanilla - CE~\cite{kamnitsas2017efficient} &  50.4 & 41.0 & 83.5 & 18.0 & 62.5 & 56.0 & 83.1 & 14.3 & 64.9 & 59.8 & 85.7 & 13.8 & 69.4 & 65.4 & 85.3 & 15.7 \\
Vanilla - CE - 80\% tumor & \cellcolor{gray!25}45.5 & \cellcolor{gray!25}36.0 & \cellcolor{gray!25}86.7 & \cellcolor{gray!25}17.8 & \cellcolor{gray!25}61.5 & \cellcolor{gray!25}54.2 & \cellcolor{gray!25}81.7 & \cellcolor{gray!25}18.5 & 65.3 & 59.6 & 85.0 & 15.1 & \cellcolor{gray!25}68.6 & \cellcolor{gray!25}64.1 & \cellcolor{gray!25}86.1 & \cellcolor{gray!25}14.8 \\
Vanilla - F1 (DSC) &  \cellcolor{gray!25}47.2 & \cellcolor{gray!25}37.4 & \cellcolor{gray!25}86.6 & \cellcolor{gray!25}15.9 & \cellcolor{gray!25}58.9 & \cellcolor{gray!25}51.1 & \cellcolor{gray!25}83.6 & \cellcolor{gray!25}20.1 & \cellcolor{gray!25}64.3 & \cellcolor{gray!25}58.1 & \cellcolor{gray!25}83.5 & \cellcolor{gray!25}16.3 & \cellcolor{gray!25}67.1 & \cellcolor{gray!25}62.5 & \cellcolor{gray!25}86.5 & \cellcolor{gray!25}15.3 \\
Vanilla - F2~\cite{hashemi2018asymmetric} & \cellcolor{gray!25}45.8 & \cellcolor{gray!25}36.9 & \cellcolor{gray!25}81.9 & \cellcolor{gray!25}17.9 & \cellcolor{gray!25}59.3 & \cellcolor{gray!25}52.2 & \cellcolor{gray!25}84.9 & \cellcolor{gray!25}18.0 & 66.4 & 61.1 & 83.4 & 14.1 & \cellcolor{gray!25}68.8 & \cellcolor{gray!25}66.0 & \cellcolor{gray!25}83.4 & \cellcolor{gray!25}13.7 \\
Vanilla - F4~\cite{hashemi2018asymmetric} & 51.6 & 42.5 & 83.8 & 18.1 & \cellcolor{gray!25}59.6 & \cellcolor{gray!25}53.0 & \cellcolor{gray!25}82.9 & \cellcolor{gray!25}18.4 & 65.9 & 61.9 & 85.4 & 14.2 & \cellcolor{gray!25}67.5 & \cellcolor{gray!25}64.5 & \cellcolor{gray!25}84.9 & \cellcolor{gray!25}13.7 \\
Vanilla - F8~\cite{hashemi2018asymmetric} & \cellcolor{gray!25}47.4 & \cellcolor{gray!25}38.7 & \cellcolor{gray!25}83.1 & \cellcolor{gray!25}19.6 & \cellcolor{gray!25}59.8 & \cellcolor{gray!25}52.4 & \cellcolor{gray!25}87.0 & \cellcolor{gray!25}15.4 & \cellcolor{gray!25}64.5 & \cellcolor{gray!25}60.3 & \cellcolor{gray!25}85.2 & \cellcolor{gray!25}14.7 & \cellcolor{gray!25}67.9 & \cellcolor{gray!25}65.4 & \cellcolor{gray!25}81.6 & \cellcolor{gray!25}14.9 \\
\hline
Large margin loss~\cite{liu2016large} & \cellcolor{gray!25}44.5 & \cellcolor{gray!25}35.9 & \cellcolor{gray!25}82.8 & \cellcolor{gray!25}20.2 & \cellcolor{gray!25}60.9 & \cellcolor{gray!25}53.5 & \cellcolor{gray!25}84.0 & \cellcolor{gray!25}17.6 & 67.0 & 61.6 & 86.1 & 14.4 & \cellcolor{gray!25}66.5 & \cellcolor{gray!25}62.2 & \cellcolor{gray!25}88.1 & \cellcolor{gray!25}13.7  \\
Asymmetric large margin loss &  56.8 & 48.9 & 83.4 & \textbf{\underline{15.0}} & 64.0 & 56.8 & 87.0 & 13.9 & 67.4 & 62.9 & 84.1 & 15.9 & \cellcolor{gray!25}68.9 & \cellcolor{gray!25}64.9 & \cellcolor{gray!25}86.5 & \cellcolor{gray!25}14.1 \\
\hline
Focal loss~\cite{lin2017focal} & 54.0 & 44.8 & 82.6 & 16.0 & 62.6 & 55.1 & 84.3 & 17.7 & 64.9 & 60.0 & 84.4 & 19.5 & \cellcolor{gray!25}67.0 & \cellcolor{gray!25}62.3 & \cellcolor{gray!25}87.0 & \cellcolor{gray!25}16.5 \\
Asymmetric focal loss & 58.8 & 51.4 & 81.6 & \textbf{\underline{15.0}} & 66.8 & 62.0 & 83.2 & \textbf{13.2} & 68.9 & 66.2 & 83.3 & \textbf{12.5} & \textbf{71.5} & 70.6 & 83.7 & 12.1 \\
\hline
Adversarial training~\cite{goodfellow2014explaining} & 53.2 & 44.6 & 85.0 & 19.2 & \cellcolor{gray!25}62.0 & \cellcolor{gray!25}55.0 & \cellcolor{gray!25}84.8 & \cellcolor{gray!25}20.6 & \cellcolor{gray!25}64.6 & \cellcolor{gray!25}59.4 & \cellcolor{gray!25}84.6 & \cellcolor{gray!25}17.3 & \cellcolor{gray!25}65.6 & \cellcolor{gray!25}61.2 & \cellcolor{gray!25}86.0 & \cellcolor{gray!25}19.4 \\
Asymmetric adversarial training & 58.5 & 50.8 & 80.1 & 16.2 & 63.9 & 58.2 & 83.1 & 17.2 & 67.7 & 63.7 & 84.2 & 17.0 & 70.5 & 68.4 & 83.0 & 14.8 \\
\hline
Mixup~\cite{zhang2017mixup} & \cellcolor{gray!25}49.7 & \cellcolor{gray!25}40.9 & \cellcolor{gray!25}83.0 & \cellcolor{gray!25}19.6 & \cellcolor{gray!25}60.3 & \cellcolor{gray!25}53.9 & \cellcolor{gray!25}83.1 & \cellcolor{gray!25}21.2 & \cellcolor{gray!25}63.9 & \cellcolor{gray!25}58.5 & \cellcolor{gray!25}84.1 & \cellcolor{gray!25}18.2 & \cellcolor{gray!25}66.4 & \cellcolor{gray!25}61.5 & \cellcolor{gray!25}86.8 & \cellcolor{gray!25}19.0 \\
Asymmetric mixup & \textbf{59.8} & 56.8 & 74.7 & 17.7 & \textbf{68.5} & 65.1 & 80.7 & 15.3 & \textbf{70.8} & 67.9 & 85.1 & \textbf{\underline{11.6}} & 70.7 & 67.9 & 85.4 & \textbf{11.8} \\
\hline
Symmetric combination & \cellcolor{gray!25}50.0 & \cellcolor{gray!25}42.0 & \cellcolor{gray!25}84.6 & \cellcolor{gray!25}21.1 & \cellcolor{gray!25}60.3 & \cellcolor{gray!25}53.1 & \cellcolor{gray!25}84.7 & \cellcolor{gray!25}25.1 & \cellcolor{gray!25}64.1 & \cellcolor{gray!25}58.3 & \cellcolor{gray!25}86.6 & \cellcolor{gray!25}19.1 & \cellcolor{gray!25}67.2 & \cellcolor{gray!25}63.1 & \cellcolor{gray!25}86.6 & \cellcolor{gray!25}15.1 \\
Asymmetric combination & \textbf{\underline{63.4}} & 63.1 & 75.9 & \textbf{15.1} & \textbf{\underline{72.4}} & 72.9 & 78.3 & \textbf{\underline{10.8}} & \textbf{\underline{71.6}} & 72.0 & 80.1 & 13.7 & \textbf{\underline{74.1}} & 76.0 & 82.4 & \textbf{\underline{10.7}} \\
\hline
\end{tabular}
\end{lrbox}
\scalebox{0.91}{\usebox{\tablebox}}
\end{table*}

\begin{table*}[t]
\centering
\caption{Evaluation of brain stroke lesion segmentation on ATLAS based on DeepMedic with different amounts of training data and different techniques to counter overfitting. The results are calculated with post-processing. Results which have worse DSC than the vanilla baseline are highlighted with shading. Best and second best results are in bold with the best also underlined.}\label{tab2}
\newsavebox{\tableboxatlas}
\begin{lrbox}{\tableboxatlas}
\begin{tabular}{p{43mm}<{\centering}|c|c|c|c||c|c|c|c||c|c|c|c}
\hlineB{3}
\multirow{2}{*}{Method} & \multicolumn{4}{c||}{30\% training} & \multicolumn{4}{c||}{50\% training} & \multicolumn{4}{c}{100\% training}  \\ 
 &  DSC & SEN &  PRC & HD &   DSC & SEN &  PRC & HD & DSC & SEN &  PRC & HD \\
\hlineB{2}

Vanilla - w/ augmentation~\cite{kamnitsas2017efficient} & 22.2 & 18.3 & 60.9 & 48.6 & 45.2 & 40.9 & 59.7 & 31.1 & 54.5 & 49.5 & 67.3 & 32.2 \\
Vanilla - w/o augmentation & 15.0 & 11.7 & 59.1 & 51.9 & 40.3 & 35.9 & 53.0 & 40.2 & 51.7 & 48.0 & 62.3 & 31.9 \\
Vanilla - asymmetric augmentation & 22.4 & 18.8 & 58.0 & 50.2 & 47.3 & 43.4 & 57.5 & 32.1 & 56.9 & 51.9 & 69.8 & 28.2 \\
\hline
Large margin loss~\cite{liu2016large} & \cellcolor{gray!25}18.9 & \cellcolor{gray!25}14.8 & \cellcolor{gray!25}64.4 & \cellcolor{gray!25}48.8 & 45.3 & 40.7 & 60.5 & 36.8 & 55.1 & 49.4 & 70.0 & 28.0 \\
Asymmetric large margin loss & 23.5 & 19.8 & 58.6 & 45.9 & 47.7 & 44.3 & 58.4 & 33.8 & \textbf{57.6} & 54.1 &67.6 & \textbf{27.8} \\
\hline
Focal loss~\cite{lin2017focal} & \cellcolor{gray!25}20.4 & \cellcolor{gray!25}16.7 & \cellcolor{gray!25}62.7 & \cellcolor{gray!25}47.9 & 46.9 & 41.8 & 61.7 & 31.4 & 56.0 & 50.8 & 69.1 & 30.9 \\
Asymmetric focal loss & 26.3 & 22.2 & 59.0 & 46.4 & 49.0 & 47.8 & 56.3 & 31.7 & 56.6 & 63.2 & 55.6 & 27.9 \\
\hline
Adversarial training~\cite{goodfellow2014explaining} & \cellcolor{gray!25}20.1 & \cellcolor{gray!25}16.7 & \cellcolor{gray!25}57.3 & \cellcolor{gray!25}56.9 & 47.2 & 41.6 & 62.6 & 35.0 & \cellcolor{gray!25}54.0 & \cellcolor{gray!25}48.5 & \cellcolor{gray!25}69.7 & \cellcolor{gray!25}34.9 \\
Asymmetric adversarial training & \textbf{28.1} & 23.3 & 63.9 & \textbf{43.4} & \textbf{50.2} & 46.5 & 64.2 & \textbf{29.6} & 55.5 & 51.6 & 69.5 & 33.6 \\
\hline
Mixup~\cite{zhang2017mixup} & \cellcolor{gray!25}14.5 & \cellcolor{gray!25}12.0 & \cellcolor{gray!25}52.7 & \cellcolor{gray!25}54.1 & 45.7 & 41.7 & 59.5 & 30.3 & \cellcolor{gray!25}53.8 & \cellcolor{gray!25}50.0 & \cellcolor{gray!25}67.8 & \cellcolor{gray!25}29.2 \\
Asymmetric mixup & 22.8 & 20.9 & 50.9 & 45.9 & 49.0 & 49.2 & 53.9 & 31.5 & 57.0 & 51.0 & 74.6 & 31.7 \\
\hline
Symmetric combination & 22.2 & 17.7 & 66.1 & 49.4 & 48.5 & 44.5 & 61.0 & 31.6 & 56.0 & 50.4 & 71.5 & 29.9 \\
Asymmetric combination & \textbf{\underline{31.1}} & 27.9 & 57.6 & \textbf{\underline{42.1}} & \textbf{\underline{52.2}} & 50.6 & 59.5 & \textbf{\underline{27.8}} & \textbf{\underline{58.5}} & 54.9 & 70.9 & \textbf{\underline{27.3}} \\
\hline
\end{tabular}
\end{lrbox}
\scalebox{1}{\usebox{\tableboxatlas}}
\end{table*}

\begin{table*}[t]
\centering
\caption{Evaluation of kidney and kidney tumor segmentation based on 3D U-Net with different amounts of training data and different techniques to counter overfitting. The results are calculated with post-processing. Results which have worse DSC than the vanilla baseline are highlighted with shading. Best and second best results are in bold with the best also underlined.}\label{tab3}
\newsavebox{\tableboxkits}
\begin{lrbox}{\tableboxkits}
\begin{tabular}{p{40mm}<{\centering}|p{4mm}<{\centering}|p{4mm}<{\centering}|p{4mm}<{\centering}|p{4mm}<{\centering}||p{4mm}<{\centering}|p{4mm}<{\centering}|p{4mm}<{\centering}|p{4mm}<{\centering}||p{4mm}<{\centering}|p{4mm}<{\centering}|p{4mm}<{\centering}|p{4mm}<{\centering}}
\hlineB{3}
\multirow{3}{*}{Method} & \multicolumn{12}{c}{Kidney} \\
 & \multicolumn{4}{c||}{10\% training} & \multicolumn{4}{c||}{50\% training} & \multicolumn{4}{c}{100\% training} \\
 & DSC & SEN &  PRC & HD &  DSC & SEN &  PRC & HD &  DSC & SEN &  PRC & HD \\
\hlineB{2}

Vanilla - w/ augmentation~\cite{isensee2019automated} & 93.3 &91.2 & 96.9 & 5.4 & 96.4 & 95.8 & 97.1 & 2.7 & 96.6 & 96.1 & 97.3 & 2.4 \\
Vanilla - w/o augmentation & 92.3 & 89.3 & 96.8 & 12.1 & 96.1 & 95.6 & 96.7 & 2.8 & 96.3 & 95.8 & 96.9 & 2.7 \\
Vanilla - asymmetric augmentation & 94.3 & 92.2 & 97.0 & 5.2 & 94.9 & 94.5 & 95.5 & 5.9 & 96.1 & 95.8 & 96.4 & 3.8 \\
\hline
Large margin loss~\cite{liu2016large} & \textbf{94.6} & 92.7 & 97.1 & 4.8 & 96.4 & 95.9 & 97.0 & 2.8 & 96.1 & 95.9 & 96.3 & 3.2 \\
Asymmetric large margin loss & 93.8 & 91.4 & 97.2 & 5.3 & 96.1 & 95.5 & 96.9 & 2.9 & \textbf{96.8} & 96.6 & 97.1 & \textbf{\underline{2.2}} \\
\hline
Focal loss~\cite{lin2017focal} & 91.4 & 85.9 & 99.2 & 10.6 & 94.1 & 89.6 & 99.2 & 4.2 & 94.3 & 90.0 & 99.1 & 4.2 \\
Asymmetric focal loss & 92.0 & 86.7 & 99.0 & 6.0 & 94.7 & 90.9 & 98.9 & 3.5 & 94.8 & 90.9 & 99.1 & 3.1 \\
\hline
Adversarial training~\cite{goodfellow2014explaining} & 94.1 & 91.9 & 97.3 & 9.1 & 96.3 & 95.7 & 97.1 & 2.6 & 96.6 & 96.2 & 97.2 & \textbf{2.3} \\ 
Asymmetric adversarial training & 94.4 & 92.5 & 97.2 & 5.7 & \textbf{96.6} & 96.0 & 97.3 & \textbf{2.5} & \textbf{96.8} & 96.4 & 97.3 & \textbf{2.3} \\
\hline
Mixup~\cite{zhang2017mixup} & \textbf{\underline{95.0}} & 93.2 & 97.3 & \textbf{\underline{4.2}} & \textbf{\underline{96.8}} & 96.2 & 97.5 & \textbf{\underline{2.3}} & \textbf{\underline{96.9}} & 96.4 & 97.5 & \textbf{\underline{2.2}}  \\
Asymmetric mixup & \textbf{94.6} & 92.6 & 97.3 & \textbf{4.5} & 96.0 & 95.2 & 97.0 & 3.1 & 96.4 & 95.7 & 97.3 & 2.7  \\
\hline
Symmetric combination &  94.1 & 91.4 & 97.5 & 5.1 & 94.6 & 91.0 & 98.7 & 4.3 & 96.7 & 96.2 & 97.2 & \textbf{\underline{2.2}} \\
Asymmetric combination & 93.5 & 89.7 & 98.5 & 5.2 & 93.9 & 90.0 & 98.3 & 5.3 & 96.7 & 95.6 & 97.9 & \textbf{\underline{2.2}} \\
\hlineB{3}
\multirow{3}{*}{Method} & \multicolumn{12}{c}{Kidney tumor} \\
 & \multicolumn{4}{c||}{10\% training} & \multicolumn{4}{c||}{50\% training} & \multicolumn{4}{c}{100\% training} \\
 & DSC & SEN &  PRC & HD &  DSC & SEN &  PRC & HD &  DSC & SEN &  PRC & HD \\
\hlineB{2}
Vanilla - w/ augmentation~\cite{isensee2019automated} & 54.6 & 46.0 & 80.0 & \textbf{53.2} & 76.0 & 72.8 & 86.1 & 25.1 & 79.2 & 77.0 & 86.2 & \textbf{17.8}  \\
Vanilla - w/o augmentation & 37.4 & 31.5 & 65.6 & 96.0 & 62.8 & 58.7 & 75.9 & 47.8 & 73.0 & 69.1 & 83.4 & 18.9  \\
Vanilla - asymmetric augmentation &  55.9 & 48.2 & 76.4 & 71.5 & \cellcolor{gray!25}74.3 & \cellcolor{gray!25}70.3 & \cellcolor{gray!25}85.2 & \cellcolor{gray!25}33.3 & \cellcolor{gray!25}78.4 & \cellcolor{gray!25}76.9 & \cellcolor{gray!25}85.7 & \cellcolor{gray!25}19.8 \\
\hline
Large margin loss~\cite{liu2016large} & \cellcolor{gray!25}52.2 & \cellcolor{gray!25}44.3 & \cellcolor{gray!25}77.2 & \cellcolor{gray!25}68.5 & 78.2 & 74.3 & 87.8 & 26.6 & 80.2 & 79.1 & 84.5 & 25.5 \\
Asymmetric large margin loss &  55.5 & 48.3 & 77.4 & 71.6 & \textbf{78.4} & 74.9 & 87.5 & 24.1 & \textbf{82.3} & 81.4 & 86.0 & \textbf{\underline{16.9}} \\
\hline
Focal loss~\cite{lin2017focal} & \cellcolor{gray!25}47.1 & \cellcolor{gray!25}37.5 & \cellcolor{gray!25}78.2 & \cellcolor{gray!25}74.5 & \cellcolor{gray!25}73.0 & \cellcolor{gray!25}66.0 & \cellcolor{gray!25}87.6 & \cellcolor{gray!25}40.2 & \cellcolor{gray!25}79.0 & \cellcolor{gray!25}73.2 & \cellcolor{gray!25}90.0 & \cellcolor{gray!25}20.3 \\
Asymmetric focal loss & \textbf{57.9} & 48.9 & 78.4 & 61.4 & 77.4 & 74.4 & 85.0 & \textbf{20.2} & 81.5 & 80.6 & 86.7 & 19.4 \\
\hline
Adversarial training~\cite{goodfellow2014explaining} & \cellcolor{gray!25}50.9 & \cellcolor{gray!25}42.5 & \cellcolor{gray!25}81.3 & \cellcolor{gray!25}62.0 & \cellcolor{gray!25}73.2 & \cellcolor{gray!25}69.6 & \cellcolor{gray!25}83.9 & \cellcolor{gray!25}44.1 & 81.9 & 81.1 & 85.8 & 27.6 \\ 
Asymmetric adversarial training & 55.2 & 47.8 & 79.6 & 66.7 & 78.3 & 74.9 & 87.9 & 23.7 & 82.1 & 81.1 & 87.4 & 19.7 \\
\hline
Mixup~\cite{zhang2017mixup} & \cellcolor{gray!25}53.3 & \cellcolor{gray!25}45.2 & \cellcolor{gray!25}81.6 & \cellcolor{gray!25}57.8 & 77.0 & 72.9 & 87.3 & 32.1  & 80.3 & 78.5 & 85.9 & 34.1 \\
Asymmetric mixup & 56.8 & 48.1 & 84.6 & 66.5 & 77.9 & 74.0 & 89.2 & 22.0 & 79.7 & 78.1 & 87.3 & 19.3 \\
\hline
Symmetric combination & \cellcolor{gray!25}53.9 & \cellcolor{gray!25}45.1 & \cellcolor{gray!25}81.3 & \cellcolor{gray!25}70.2 & \cellcolor{gray!25}73.9 & \cellcolor{gray!25}67.1 & \cellcolor{gray!25}87.7 & \cellcolor{gray!25}39.6 & 80.9 & 79.3 & 86.5 & 19.6 \\
Asymmetric combination & \textbf{\underline{59.2}} & 52.2 & 80.3 & \textbf{\underline{49.5}} & \textbf{\underline{79.4}} & 77.0 & 86.7 & \textbf{\underline{15.5}} & \textbf{\underline{82.7}} & 82.1 & 87.0 & 18.8  \\
\hline
\end{tabular}
\end{lrbox}
\scalebox{1}{\usebox{\tableboxkits}}
\end{table*}

\subsubsection{Kidney tumor segmentation}

In addition, we evaluate the asymmetric techniques for the case of kidney tumor segmentation. We train a well configured 3D U-Net \cite{isensee2019automated} which includes extensive data augmentation with scaling, rotations, brightness, contrast, gamma and Gaussian noise augmentations with a predefined policy \cite{heller2019kits19}. We also train DeepMedic with similar augmentation strategies yielding lower accuracy on this task. Therefore, we evaluate the asymmetric regularization techniques on the U-Net with kidney tumor segmentation. The task includes the segmentation of both the kidney and kidney tumor. As the segmentation of kidney is relatively easy, in this experiment we only focus on tumor segmentation and only take kidney tumor as the foreground class to implement asymmetric techniques. To be specific, we always set $\boldsymbol{r}$ as ${\left[0,0,1\right]}^\intercal$. The network is always trained with both CE and sample-wise DSC loss. The two losses have the same weight. We conduct experiments using the training dataset of KiTS19 dataset~\cite{heller2019kits19} which contains 210 CT images. We resample all the CT images to a common voxel spacing of 1.6$\times$1.6$\times$3.2 mm following~\cite{isensee2019automated}. The original and asymmetric techniques are implemented on both loss functions with the same hyper-parameters. We test on 70 cases and train separate models using 140 (100\%), 70 (50\%), 28 (20\%), 14 (10\%) and 7 cases (5\% of full training set). Note, we do not manually clean the training dataset or employ deep supervision as done in~\cite{isensee2019automated}, therefore the absolute performance of our U-Net is not directly comparable with the reported challenge results.

\subsection{Quantitative results}

Taking the provided manual segmentations as the ground truth, we calculate DSC, sensitivity (SEN), precision (PRC) and 95\% Hausdorff distance (HD) (mm) to evaluate the segmentation accuracy. The initial segmentation results of our method always have higher sensitivity and DSC, but sometimes would cause more false positive predictions and therefore lead to worse distance-based metrics such as HD. We argue that in practice this problem can be addressed by taking advantage of some connected component-based post-processing, which is widely adopted in many segmentation methods~\cite{isensee2019automated}. Specifically, we assume there is only one target component and suppress all but the largest region. We report both results with or without these post-processing operations. The quantitative segmentation results on BRATS, ATLAS and KiTS datasets using different amounts of training data with post-processing are summarized in Table~\ref{tab1}, \ref{tab2} and \ref{tab3}, respectively. The corresponding quantitative segmentation results without post-processing are summarized in Supplementary Table~\ref{taba4}, \ref{taba5} and \ref{taba6}. We also evaluate one of the proposed methods, asymmetric focal loss, with abdominal organ segmentation in which multiple classes are considered under-represented. The experiments are summarized in Supplementary Table~\ref{taba8}.

Class imbalance affects the segmentation sensitivity of the under-represented class, as shown in Section~\ref{sec:motivation}. We find that previous attempts to tackle class imbalance do not improve sensitivity, while our asymmetric methods do lead to better results with higher sensitivity across different tasks. This indicates that the proposed methods may effectively mitigate overfitting under class imbalance. These results in turn support our previous analysis of logit shift under class imbalance and indicate that considering this implication would help build unbiased network.

\subsubsection{Baseline experiments}

\begin{figure*}[t]
\centering
\includegraphics[width=\textwidth]{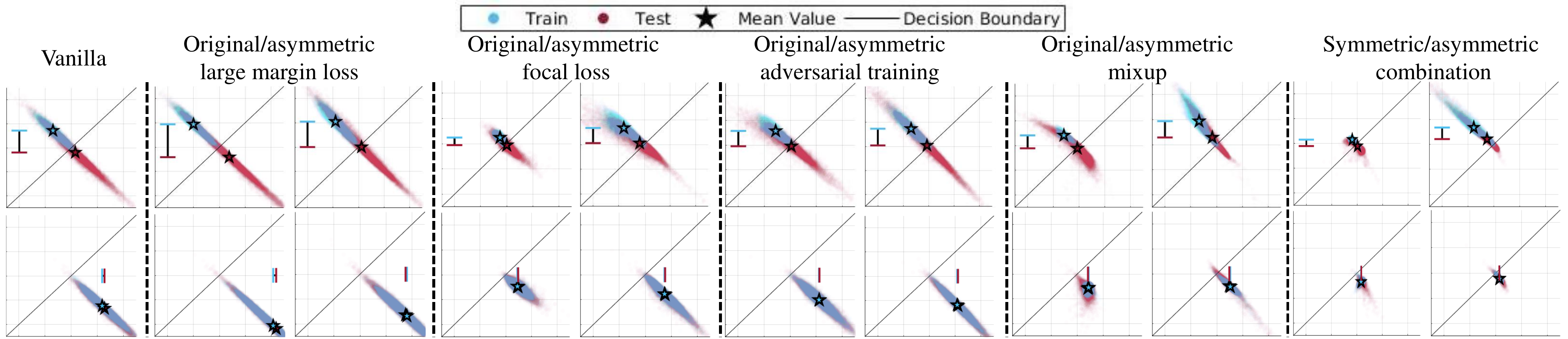}

\caption{Activations of the classification layer when processing tumor (top) and background (bottom) samples of BRATS with DeepMedic, using 5\% training data. Asymmetric modifications lead to better separation of the logits of unseen tumor samples.} \label{fig8}
\end{figure*}

\begin{figure*}[t]
\centering
\includegraphics[width=\textwidth]{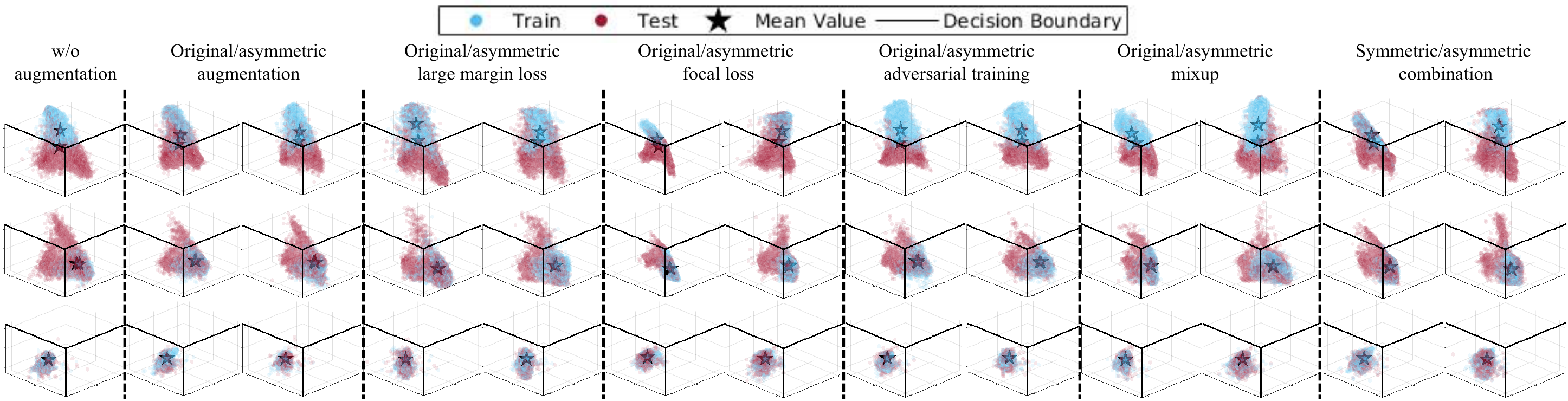}

\caption{Activations of the classification layer for tumor (top), kidney (middle) and background (bottom) samples of KiTS with 3D U-Net, using 10\% training data. Asymmetric modifications also lead to better separation of the logits of unseen tumor samples and is complementary to standard data augmentation.} \label{fig9}
\end{figure*}

We perform baseline experiments with binary brain tumor core segmentation, as shown in Table~\ref{tab1}. We show that increasing the weight of tumor samples (from 50\% to 80\%) decreases performance when the dataset is highly imbalanced. This is because increasing the weight encourages the network to memorize the under-represented samples and may actually lead to more overfitting, thus being counter-productive. Simply changing the objective function to F-score (defined as $ F_{\beta}=(1+\beta^2)\frac{\text{sensitivity} \cdot \text{precision}}{\beta^2\text{sensitivity} + \text{precision}}$), which is a balancing loss and weights sensitivity $\beta$-times more than precision~\cite{milletari2016v,hashemi2018asymmetric}, only shows little improvements increasing the sensitivity slightly. Changing the sampling weights or training with a loss function using F-scores seems to have little impact when foreground training samples are limited, with training accuracy close to 100\%, as shown in Fig.~\ref{fig1}.

A common approach to alleviate under-segmentation is to adjust thresholds of decision boundaries based on validation sets. In this work, however, we observe distribution shift on unseen test data, which can significantly differ from the training/validation sets. Hence, a threshold selected on validation data may not be optimal for new test data. Due to the lack of ground truth, it is practically not possible to optimise the decision thresholds for a specific test set.

\subsubsection{Asymmetric large margin loss}

The original large margin loss decreases performance in some cases, while our modification yields improvements over the symmetric version in all cases.

\subsubsection{Asymmetric focal loss}

The original focal loss also decreases the sensitivity in some cases and leads to worse performance. It is because the focal term would decrease the weight of foreground samples and push its logit closer to the decision boundary, making it easier to cause false negative predictions. Our modification removes the loss attenuation for the under-represented class and improve the performance in all cases. We notice that the asymmetric focal loss would make the performance of other class (kidney) overfit more, but it is not the focus of this study and can be easily addressed by just keeping focal term for the background class.

\subsubsection{Asymmetric adversarial training}

When the network is trained without data augmentation (as shown in Table~\ref{tab1}), the original adversarial training seems to be effective when little training data is available while our modifications can further improve the sensitivity and boost the performance substantially. 

When the network is trained with data augmentation (as shown in Table~\ref{tab2} and \ref{tab3}), we find the original adversarial training does not improve the performance when training data is limited. It indicates that in this case the augmented samples by adversarial training might not add anything on top of intensity augmentation. In contrast, our proposed modifications seem to help always leading to better segmentation performance.

\subsubsection{Asymmetric mixup}

We find the original mixup can be effective for the well-represented class. For example, it always improve the segmentation performance of kidney, as shown in Table~\ref{tab3}. However, the original mixup leads to lower sensitivity for the under-represented class. We find that the asymmetric mixup can improve the performance for BRATS to a large extent, as shown in Table~\ref{tab1}. However, we also find it to be less effective for ATLAS and KiTS which only have one image channel, as shown in Table~\ref{tab2} and \ref{tab3}. This may be because the intensity distributions of healthy and lesion regions in ATLAS and KiTS overlap too much, as shown in the Supplementary Fig.~\ref{figa7}. In this case, the mixed samples, which are very similar to the background samples in intensity but are taken as the foreground samples, may confuse the network. For BRATS, however, four image channels are available with the healthy and tumor regions show larger differences in T2 and fluid-attenuated inversion recovery (FLAIR) sequences. The mixed samples seem to take good advantage of the intensity relationship.

\subsubsection{Asymmetric augmentation}

Despite its simplicity, we find asymmetric augmentation to be an effective method to improve segmentation performance of the under-represented classes in most cases in terms of DSC and sensitivity, as summarized in Table~\ref{tab2}. However, we also notice that it could decrease the performance when data augmentation is strong and training data is sufficient, as shown in Table~\ref{tab3}. It might be because the strong asymmetric augmentation would drive the model to focus too much on the foreground samples, making the background samples under-represented.

\subsubsection{The combination of asymmetric techniques}

We also combine the asymmetric techniques which are found to improve the segmentation accuracy. The combination of the asymmetric techniques is a safe choice leading to the overall best segmentation results with improved sensitivity in all cases. In contrast, the combination of the symmetric counterparts is unable to mitigate overfitting and often decreases sensitivity. 

\subsection{Logit distribution changes}

The effects of all the techniques on the logit distributions of BRATS using 5\% training data is presented in Fig.~\ref{fig8}. Asymmetric techniques would increase the variances of the foreground class and expand its logit distribution. The original large margin loss and adversarial training try to push samples from different classes far from each other, however, the logits of unseen data remain in the center around the decision boundary and thus the predictions are not improved. The original large margin loss results in even larger shifts for the foreground samples. For our asymmetric modifications only the logits of foreground samples are pushed away and the unseen foreground logits tend to remain on the correct side of the decision boundary. The original focal loss encourages the network to prevent the logits of each class from staying too far from the decision boundary. However, it allows foreground logits to remain near the decision boundary which can result in false negative predictions on unseen samples. Our asymmetric focal loss removes the constraints for foreground samples. Original mixup encourages the symmetric distributions of different classes but does not consider class imbalance. Asymmetric mixup exploits the latent space based on the relationship between samples to generate foreground samples and make the decision boundary stay near the background class. This leads to the largest improvement by increasing the region for the foreground logit distribution and reduce logit shift of unseen foreground samples. The combination of the four asymmetric techniques can stabilize the logits further more. 

The effects of all the techniques on the logit distribution of KiTS using 10\% training data is summarized in Fig.~\ref{fig9}. The original data augmentation can reduce the logit shift of both unseen kidney and kidney tumor samples although it is not specifically designed to regularize the logit distribution. The asymmetric augmentation can further reduce the logit shift of the unseen tumor samples. The asymmetric large margin loss reduces the tumor logit shift towards the kidney class. Although the logit distribution is already regularized by the strong augmentation, the proposed asymmetric techniques provide further benefits in stabilizing the logits.

\section{Conclusion}
\label{sec5}

We study overfitting of neural networks under class imbalance by inspecting network behavior. We observe that when processing unseen under-represented samples, the logit activations tend to shift towards the decision boundary and the sensitivity decreases. This phenomenon is confirmed across a variety of different tasks and two popular different network architectures. We derive simple yet effective asymmetric variants of existing loss functions and regularization techniques to prevent overfitting. We show that our proposed methods can substantially improve segmentation performance under class imbalance in terms of DSC and increased sensitivity, outperforming previous solutions. We believe more regularization methods can be derived to alleviate this problem by considering the biased network behavior. We also believe that the plotting logit distributions may be useful network inspection tool and help to gain a better understanding network behavior under different training scenarios. In future work, we will investigate if monitoring of intermediate activations may provide further insights for other challenging settings such as domain shift or self-supervised learning.



\section*{Acknowledgements}
This work received funding from the European Research Council (ERC) under the European Union's Horizon 2020 research and innovation programme (grant agreement No 757173, project MIRA, ERC-2017-STG). ZL is supported by the China Scholarship Council (CSC). KK is funded by the UKRI London Medical Imaging \& Artificial Intelligence Centre for Value Based Healthcare.

\bibliographystyle{ieee}
\bibliography{reference}

\section*{Supplementary Material}

\section{The analysis of large margin loss from a sample re-weighting perspective}
\label{sec:marginlossanalysis}

Some of the proposed methods are based on two well known loss functions, noted as focal loss and large margin loss. We argue that these two loss functions can be both seen as sample-level re-weighting methods, which change the magnitude of gradient of the network output by multiplying a scalar. Specifically, focal loss would decrease the weights of well-classified samples, while large margin loss would increase the weights of all samples, especially for well-classified samples. Here, we provide an analysis of these loss functions from sample re-weighting perspective. 

Following the formulation in the main text, we first consider a CNN trained using cross-entropy loss with one sample $\boldsymbol{x_{i}}$ and its one-hot label $\boldsymbol{y_{i}}$:

\begin{equation}
    L_{CE}(\boldsymbol{x_{i}},\boldsymbol{y_{i}}) = -\sum_{j=1}^{c} y_{ij}\log(p_{ij}) = -\boldsymbol{y_{i}}\cdot\log(\boldsymbol{p_{i}}),
    \label{eq:CEa}
\end{equation}

where $\boldsymbol{p_{i}}$ is the calculated probability, which is a normalized term from the network output $\boldsymbol{z_{i}}$:

\begin{equation}
    \boldsymbol{p_{i}} = \frac{\mathrm{e}^{\boldsymbol{z_{i}}}}{\mathrm{e}^{\boldsymbol{z_{i}}}\cdot\boldsymbol{1}}.
    \label{eq:proba}
\end{equation}

Let's look at the gradient of a general loss function $\emph{L}(\boldsymbol{x_{i}},\boldsymbol{y_{i}})$ with respect to the network parameters $\theta$:

\begin{equation}
    \frac{\partial L(\boldsymbol{x_{i}},\boldsymbol{y_{i}})}{\partial \theta} = \frac{\partial L(\boldsymbol{x_{i}},\boldsymbol{y_{i}})}{\partial \boldsymbol{z_{i}}} \frac{\partial \boldsymbol{z_{i}}}{\partial \theta},
    \label{eq:gradtheta}
\end{equation}

where the former term is associated with the design of loss functions and the latter is related to the network architecture. Considering a cross entropy loss $L_{CE}(\boldsymbol{x_{i}},\boldsymbol{y_{i}})$,  we can have:

\begin{equation}
\resizebox{0.6\hsize}{!}{$
\begin{aligned}
    & \frac{\partial L_{CE}(\boldsymbol{x_{i}},\boldsymbol{y_{i}})}{\partial \boldsymbol{z_{i}}} = \frac{\partial L_{CE}(\boldsymbol{x_{i}},\boldsymbol{y_{i}})}{\partial \boldsymbol{p_{i}}}\frac{\partial \boldsymbol{p_{i}}}{\partial \boldsymbol{z_{i}}} \\
    & = -\frac{1}{\boldsymbol{p_{i}}\cdot\boldsymbol{y_{i}}}\boldsymbol{p_{i}}\cdot\boldsymbol{y_{i}}(\boldsymbol{y_{i}} - \boldsymbol{p_{i}}) = \boldsymbol{p_{i}} - \boldsymbol{y_{i}}.
    \label{eq:grad}
\end{aligned}$}
\end{equation}

In instance-level, we can assign different weights for different samples $\boldsymbol{x_{i}}$ with different scalars $w_i$. The weights could be derive based on the frequency of samples or other heuristic rules. Typically, we multiply $L_{CE}(\boldsymbol{x_{i}},\boldsymbol{y_{i}})$ by $w_i$, and the gradient after re-weighting would become:

\begin{equation}
    \frac{\partial \Big(w_iL_{CE}(\boldsymbol{x_{i}},\boldsymbol{y_{i}})\Big)}{\partial \boldsymbol{z_{i}}} = w_i(\boldsymbol{p_{i}} - \boldsymbol{y_{i}}).
    \label{eq:gradreweight}
\end{equation}

It can be seen that re-weighting would change the gradient of the network output for different samples and make the model fit better the chosen samples, which are assigned a larger weight $w_i$.

Similarly, we can also derive the gradient of the focal loss as:

\begin{equation}
\resizebox{1\hsize}{!}{$
\begin{aligned}
    & \frac{\partial \Big(L_{CE_{focal}}(\boldsymbol{x_{i}},\boldsymbol{y_{i}})\Big)}{\partial \boldsymbol{z_{i}}} = \frac{\partial \Big(L_{CE_{focal}}(\boldsymbol{x_{i}},\boldsymbol{y_{i}})\Big)}{\partial \boldsymbol{p_{i}}}\frac{\partial \boldsymbol{p_{i}}}{\partial \boldsymbol{z_{i}}} \\
    & = \Big((1-\boldsymbol{p_{i}}\cdot\boldsymbol{y_{i}})^{\gamma}-\gamma \boldsymbol{p_{i}}\cdot\boldsymbol{y_{i}}\log(\boldsymbol{p_{i}}\cdot\boldsymbol{y_{i}})(1-\boldsymbol{p_{i}}\cdot\boldsymbol{y_{i}})^{\gamma-1}\Big)(\boldsymbol{p_{i}} - \boldsymbol{y_{i}}) \\
    & = {w_{i_{focal}}}(\boldsymbol{p_{i}} - \boldsymbol{y_{i}}),
    \label{eq:gradfocalloss}
\end{aligned}$}
\end{equation}

The weight term of focal loss $w_{i_{focal}}$ is a scalar and related to the sample probability $\boldsymbol{p_{i}}\cdot\boldsymbol{y_{i}}$. Generally speaking, $w_{i_{focal}}$ would decrease when $\boldsymbol{p_{i}}\cdot\boldsymbol{y_{i}}$ is large, therefore focal loss would make the model fit the easy cases less.

More interestingly, we next look into the effect of large margin loss on the gradient. Large margin loss would change the calculation of probability for the training process. Specifically, we substitute $\boldsymbol{p_{i}}$ with $\boldsymbol{q_{i}}$ to calculate the loss function, where we require:

\begin{equation}
    \boldsymbol{q_{i}} = \frac{\mathrm{e}^{\boldsymbol{z_{i}} - \boldsymbol{y_{i}}m}}{\mathrm{e}^{\boldsymbol{z_{i}} - \boldsymbol{y_{i}}m}\cdot\boldsymbol{1}},
    \label{eq:marginproba}
\end{equation}

where $\emph{m}$ is the hyper-parameter for the margin. In this case, the gradient of large margin loss can be derived as:

\begin{equation}
\resizebox{0.7\hsize}{!}{$
\begin{aligned}
    & \frac{\partial L_{CE_M}(\boldsymbol{x_{i}},\boldsymbol{y_{i}})}{\partial \boldsymbol{z_{i}}} = \frac{\partial L_{CE}(\boldsymbol{x_{i}},\boldsymbol{y_{i}})}{\partial \boldsymbol{q_{i}}}\frac{\partial \boldsymbol{q_{i}}}{\partial \boldsymbol{z_{i}}} \\
    & = \frac{\mathrm{e}^{\boldsymbol{z_{i}}}\cdot \boldsymbol{1}}{\mathrm{e}^{\boldsymbol{z_{i}} - \boldsymbol{y_{i}}m}\cdot\boldsymbol{1}}(\boldsymbol{p_{i}} - \boldsymbol{y_{i}}) = w_{i_{M}}(\boldsymbol{p_{i}} - \boldsymbol{y_{i}}).
    \label{eq:gradmargin}
\end{aligned}$}
\end{equation}

The weight term of large margin loss $w_{i_{M}}$ is also a scalar and related to the network output $\boldsymbol{z_{i}}\cdot\boldsymbol{y_{i}}$. It can be seen that the existence of a margin $\emph{m}$ would increase the gradient of sample $\boldsymbol{x_{i}}$. Moreover, $w_{i_{M}}$ would be larger as $\boldsymbol{z_{i}} \cdot \boldsymbol{y_{i}}$ becomes larger, therefore large margin loss would make the model fit the easy cases more, and keep the distribution of $\boldsymbol{z_{i}}$ away from the decision boundary.

The analysis for $L_{DSC}$ can be done in a similar way.

\section{The magnitude of focal DSC loss}

\begin{figure}[ht]
\centering
\includegraphics[width=0.44\textwidth]{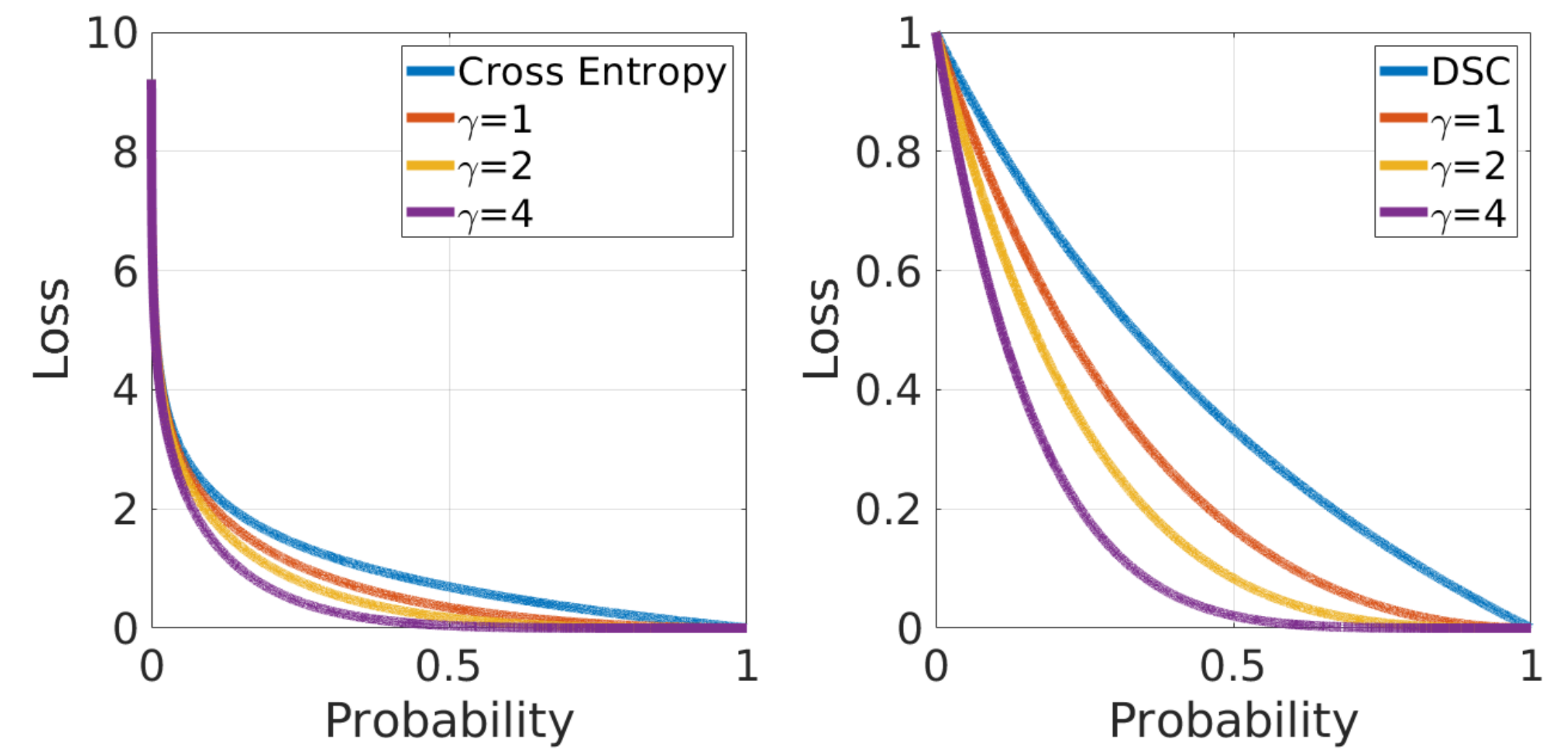}
\caption{The comparison of focal loss with cross entropy and DSC loss. The behavior of focal loss for cross entropy and DSC loss are similar using the formulation in equation \ref{eq:Focal} and \ref{eq:DSCfocal}. } \label{figa6}
\end{figure}

The proposed focal DSC loss has similar behaviour with the original of focal loss, as shown in Figure \ref{figa6}. In addition, it does not change the magnitude of loss too much compared with existing solutions~\cite{abraham2019novel, wong20183d}, making it easier to be combined with other losses. We find it is particularly important for our experiments with 3D U-net~\cite{isensee2019automated} because this framework adopts a loss function which is a combination of cross entropy and DSC loss.

\section{Hyper-parameters of the regularization techniques}

We summarize the hyper-parameters in Table~\ref{taba1}, Table~\ref{taba2} and Table~\ref{taba3} as a reference for practitioners. We find when the asymmetric regularization techniques are combined together, the network could be regularized too much. In this case, the model would not converge and even perform poorly on the training data. Therefore, we always choose hyper-parameters with smaller regularization magnitude for the experiments with the combined regularization. Empirically, we find decreasing the hyper-parameters of large margin loss and/or focal loss is a sensible choice. 

\begin{table}[ht]
\centering
\caption{Hyper-parameters of experiments using DeepMedic with BRATS.}
\label{taba1}
\begin{lrbox}{\tableboxh}
\begin{tabular}{c|c|c|c|c|c}
\hlineB{3}
\multicolumn{2}{c|}{BRATS} &  5\% data & 10\% data & 20\% data &  50\% data \\
\hlineB{2}
\multirow{7}*{Individual} & large margin \emph{m} &  1 & 0.2 & 1 & 1 \\
& focal $\gamma$ &  2 & 4 & 4 & 4 \\
& adversarial $\epsilon$ &  1e-5 & 1e-5 & 1e-5 & 1e-5 \\
& adversarial \emph{l} & 10 & 10 & 10 & 20 \\
& mixup $\lambda$ (symmetric) & 0.2 & 0.2 & 0.2 & 0.2  \\
& mixup $\lambda$ (asymmetric) & 1 & 1 & 1 & 1   \\
& mixup \emph{m} &  0.2 & 0.2 & 0.2 & 0.1 \\
\hline
\multirow{7}*{Combination} & large margin \emph{m} & 1 & 0 & 0 & 0 \\
& focal $\gamma$ & 1.5 & 2 & 2 & 4 \\
& adversarial $\epsilon$ & 1e-5 & 1e-5 & 1e-5 & 1e-5 \\
& adversarial \emph{l} & 10 & 10 & 10 & 20 \\
& mixup $\lambda$ (symmetric) & 0.2 & 0.2 & 0.2 & 0.2  \\
& mixup $\lambda$ (asymmetric) & 1 & 1 & 1 & 1 \\
& mixup \emph{m} &  0.2 & 0.2 & 0.2 & 0.1 \\
\hline
\end{tabular}
\end{lrbox}
\scalebox{0.8}{\usebox{\tableboxh}}
\end{table}

\begin{table}[ht]
\centering
\caption{Hyper-parameters of experiments using DeepMedic with ATLAS.}
\label{taba2}
\newsavebox{\tableboxa}
\begin{lrbox}{\tableboxa}
\begin{tabular}{c|m{30mm}<{\centering}|c|c|c}
\hlineB{3}
\multicolumn{2}{c|}{ATLAS} &  30\% data & 50\% data &  100\% data \\
\hlineB{2}
\multirow{9}*{Individual} & large margin \emph{m} & 0.1 & 3 & 2 \\
& focal $\gamma$ & 4 & 4 & 4 \\
& adversarial $\epsilon$ & 1e-5 & 1e-5 & 1e-5 \\
& adversarial \emph{l} & 10  & 10 & 10 \\
& mixup $\lambda$ (symmetric) & 0.2  & 0.2 & 0.2 \\
& mixup $\lambda$ (asymmetric) & 1 & 1 & 1 \\
& mixup \emph{m} & 0.8 & 0.8 & 0.2 \\
& Probability of background samples being augmented & 50\% & 50\% & 25\% \\
\hline
\multirow{9}*{Combination} & large margin \emph{m} & 0.1 & 0 & 1  \\
& focal $\gamma$ & 4 & 3 & 2  \\
& adversarial $\epsilon$ & 1e-5  & 1e-5 & 1e-5 \\
& adversarial \emph{l} & 10 & 10 & 10 \\
& mixup $\lambda$ (symmetric) & 0.2 & 0.2 & 0.2  \\
& mixup $\lambda$ (asymmetric) & ------ & ------ & 1 \\
& mixup \emph{m} & ------& ------ & 0.2 \\
& Probability of background samples being augmented & 50\% & 50\% & ------ \\
\hline
\end{tabular}
\end{lrbox}
\scalebox{0.8}{\usebox{\tableboxa}}
\end{table}

\begin{table}[ht]
\centering
\caption{Hyper-parameters of experiments using 3D U-Net with KiTS.}
\label{taba3}
\newsavebox{\tableboxhk}
\begin{lrbox}{\tableboxhk}
\begin{tabular}{c|m{30mm}<{\centering}|c|c|c}
\hlineB{3}
\multicolumn{2}{c|}{KiTS} & 10\% data & 50\% data &  100\% data \\
\hlineB{2}
\multirow{9}*{Individual} & large margin \emph{m} & 0.8 & 0.6 & 0.8 \\
& focal $\gamma$ & 6 & 6 & 6  \\
& adversarial $\epsilon$ & 1e-5 & 1e-5 & 1e-5  \\
& adversarial \emph{l} & 100 & 50 & 50 \\
& mixup $\lambda$ (symmetric) & 0.2 & 0.2 & 0.2  \\
& mixup $\lambda$ (asymmetric) & 1 & 1 & 1  \\
& mixup \emph{m} & 0.05 & 0.05 & 0.2  \\
& Probability of background samples being augmented & 0\% & 0\% & 50\% \\
\hline
\multirow{9}*{Combination} & large margin \emph{m} & 0.8 & 0.2 & 0.8  \\
& focal $\gamma$ & 4 & 6 & 2  \\
& adversarial $\epsilon$  & ------ & ------ & ------ \\
& adversarial \emph{l}& ------ & ------ & ------ \\
& mixup $\lambda$ (symmetric) & ------ & ------ & ------ \\
& mixup $\lambda$ (asymmetric) & ------ & ------ & ------ \\
& mixup \emph{m} & ------ & ------ & ------ \\
& Probability of background samples being augmented & 0\% & ------ & ------ \\
\hline
\end{tabular}
\end{lrbox}
\scalebox{0.8}{\usebox{\tableboxhk}}
\end{table}

\section{Sensitivity analysis}

We conduct a series of controlled experiments with different hyper-parameters to provide more practical details of the proposed regularization techniques. Specifically, we use a baseline DeepMedic model for brain tumor core segmentation with 5\% training data of BRATS. The experimental details are consistent with descriptions in Section~\ref{sec4}. We summarize the results with and without any post-processing in Table~\ref{taba7}. We can see from the results that the proposed methods can improve the baseline segmentation results with varied hyper-parameters in most cases. Specifically, asymmetric large margin loss yields improvements for most cases, however, a specific hyper-parameter may yield unexpected results (i.e. $\emph{m}=0.5$). A potential reason is that the model which focuses on a small portion of easy under-represented samples (c.f. equation~\ref{eq:gradmargin} in Section~\ref{sec:marginlossanalysis}) would overfit more. Asymmetric large margin loss with larger $\emph{m}$ makes the model emphasize on more under-represented samples and therefore generalize better. Asymmetric adversarial training and asymmetric mixup yields considerable improvements when the perturbation in data augmentation is larger (i.e. $\emph{l}>2.5$ for asymmetric adversarial training and $\emph{m}<0.8$ for asymmetric mixup). Asymmetric focal loss is robust and can improve the segmentation results with all chosen hyper-parameters. Therefore, we recommend to choose asymmetric focal loss at first for new applications.

\begin{table}[t]
\centering
\caption{The sensitivity analysis of different hyper-parameters. We conduct experiments with different parameters with brain tumor core segmentation (5\% training data) with BRATS using DeepMedic. Results which have worse DSC than the vanilla baseline are highlighted with gray shading.}\label{taba7}
\newsavebox{\tableboxsens}
\begin{lrbox}{\tableboxsens}
\begin{tabular}{p{20mm}<{\centering}|c|c|c|c|c|c}
\hlineB{3}
Method & \multicolumn{2}{c|}{Parameter} &  DSC & SEN &  PRC & HD \\
\hlineB{2}
\multicolumn{7}{c}{w/ post-processing} \\
\hline
Vanilla - CE  & \multicolumn{2}{c|}{------} & 50.4 & 41.0 & 83.5 & 18.0 \\
\hline
\multirow{6}{*}{\parbox{2cm}{\centering Asymmetric large margin loss}} & \multicolumn{2}{c|}{m = 0.2} & 53.6 & 44.8 & 84.8 & 15.6 \\
& \multicolumn{2}{c|}{m = 0.5} & \cellcolor{gray!25}48.4 & \cellcolor{gray!25}39.4 & \cellcolor{gray!25}81.5 & \cellcolor{gray!25}16.9 \\
& \multicolumn{2}{c|}{m = 1} & 56.8 & 48.9 & 83.4 & 15.0 \\
& \multicolumn{2}{c|}{m = 1.5} & 54.1 & 45.6 & 81.7 & 15.3  \\
& \multicolumn{2}{c|}{m = 2} & 51.6 & 42.8 & 84.0 & 16.7  \\
& \multicolumn{2}{c|}{m = 3} &  54.4 & 45.7 & 82.3 & 14.3 \\
\hline
\multirow{6}{*}{\parbox{2cm}{\centering Asymmetric focal loss}} & \multicolumn{2}{c|}{$\gamma$ = 0.5} & 53.4 & 44.8 & 79.6 & 16.6 \\
& \multicolumn{2}{c|}{$\gamma$ = 1} & 53.9 & 45.2 & 81.9 & 17.9 \\
& \multicolumn{2}{c|}{$\gamma$ = 1.5} & 56.5 & 48.3 & 87.8 & 13.8  \\
& \multicolumn{2}{c|}{$\gamma$ = 2} & 58.8 & 51.4 & 81.6 & 15.0  \\
& \multicolumn{2}{c|}{$\gamma$ = 3} & 57.5 & 49.0 & 85.5 & 14.2  \\
& \multicolumn{2}{c|}{$\gamma$ = 4} & 55.2 & 48.5 & 78.3 & 15.5  \\
\hline
\multirow{7}{*}{\parbox{2cm}{\centering Asymmetric adversarial training}} &  $\epsilon$ = 1e-5 & \emph{l} = 2.5 & \cellcolor{gray!25}50.3 & \cellcolor{gray!25}41.8 & \cellcolor{gray!25}82.0 & \cellcolor{gray!25}17.2 \\
& $\epsilon$ = 1e-5 & \emph{l} = 5 & 58.1 & 50.0 & 84.7 & 14.1 \\
& $\epsilon$ = 1e-5 & \emph{l} = 10 & 58.5 & 50.8 & 80.1 & 16.2 \\
& $\epsilon$ = 1e-5 & \emph{l} = 15 & 53.8 & 46.2 & 80.1 & 16.9\\
& $\epsilon$ = 1e-5 & \emph{l} = 20 & 56.6 & 50.7 & 76.9 & 18.8 \\
\cline{2-7}
& $\epsilon$ = 1e-4 & \emph{l} = 10 & 57.6 & 51.1 & 78.9 & 16.1 \\
& $\epsilon$ = 1e-6 & \emph{l} = 10 & 56.2 & 48.5 & 81.1 & 17.8 \\
\hline
\multirow{7}{*}{\parbox{2cm}{\centering Asymmetric mixup}} & \multicolumn{2}{c|}{m = 0.1} & 52.1 & 47.3 & 73.8 & 20.7 \\
& \multicolumn{2}{c|}{m = 0.15} & 58.1 & 53.7 & 75.0 & 19.9 \\
& \multicolumn{2}{c|}{m = 0.2} & 59.8 & 56.8 & 74.7 & 17.7 \\
& \multicolumn{2}{c|}{m = 0.25} & 60.4 & 55.0 & 82.0 & 15.6 \\
& \multicolumn{2}{c|}{m = 0.3} & 59.1 & 54.3 & 82.0 & 15.3 \\
& \multicolumn{2}{c|}{m = 0.4} & 52.1 & 44.2 & 84.2 & 21.4 \\
& \multicolumn{2}{c|}{m = 0.8} & \cellcolor{gray!25}50.3 & \cellcolor{gray!25}41.6 & \cellcolor{gray!25}85.5 & \cellcolor{gray!25}17.7 \\
\hlineB{2}
\multicolumn{7}{c}{w/o post-processing} \\
\hline
Vanilla - CE  & \multicolumn{2}{c|}{------} & 51.0 & 42.6 & 78.6 & 17.5 \\
\hline
\multirow{6}{*}{\parbox{2cm}{\centering Asymmetric large margin loss}} & \multicolumn{2}{c|}{m = 0.2} & 53.2 & 46.0 & 79.8 & 18.3 \\
& \multicolumn{2}{c|}{m = 0.5} & \cellcolor{gray!25}48.8 & \cellcolor{gray!25}40.8 & \cellcolor{gray!25}78.1 & \cellcolor{gray!25}17.5 \\
& \multicolumn{2}{c|}{m = 1} & 55.5 & 50.6 & 76.2 & 23.9 \\
& \multicolumn{2}{c|}{m = 1.5} & 52.6 & 47.2 & 73.1 & 25.8 \\
& \multicolumn{2}{c|}{m = 2} & 51.4 & 44.2 & 78.2 & 18.8 \\
& \multicolumn{2}{c|}{m = 3} & 53.4 & 47.3 & 75.1 & 21.3 \\
\hline
\multirow{6}{*}{\parbox{2cm}{\centering Asymmetric focal loss}} & \multicolumn{2}{c|}{$\gamma$ = 0.5} & 54.2 & 48.0 & 76.2 & 22.0 \\
& \multicolumn{2}{c|}{$\gamma$ = 1} & 53.7 & 46.8 & 76.0 & 22.8 \\
& \multicolumn{2}{c|}{$\gamma$ = 1.5} & 54.3 & 49.6 & 76.3 & 25.9 \\
& \multicolumn{2}{c|}{$\gamma$ = 2} & 57.3 & 52.7 & 76.4 & 24.4  \\
& \multicolumn{2}{c|}{$\gamma$ = 3} &  55.7 & 50.3 & 75.4 & 24.6 \\
& \multicolumn{2}{c|}{$\gamma$ = 4} & 54.4 & 50.3 & 71.1 & 25.4 \\
\hline
\multirow{7}{*}{\parbox{2cm}{\centering Asymmetric adversarial training}} &  $\epsilon$ = 1e-5 & \emph{l} = 2.5 & \cellcolor{gray!25}50.5 & \cellcolor{gray!25}43.6 & \cellcolor{gray!25}76.3 & \cellcolor{gray!25}21.3 \\
& $\epsilon$ = 1e-5 & \emph{l} = 5 & 56.6 & 51.3 & 76.1 & 21.9 \\
& $\epsilon$ = 1e-5 & \emph{l} = 10 & 56.8 & 51.8 & 74.8 & 23.6 \\
& $\epsilon$ = 1e-5 & \emph{l} = 15 & 53.3 & 47.6 & 74.8 & 21.5 \\
& $\epsilon$ = 1e-5 & \emph{l} = 20 & 55.4 & 53.2 & 72.0 & 26.2 \\
\cline{2-7}
& $\epsilon$ = 1e-4 & \emph{l} = 10 & 56.9 & 53.3 & 74.1 & 22.4 \\
& $\epsilon$ = 1e-6 & \emph{l} = 10 & 55.2 & 50.0 & 76.0 & 23.8 \\
\hline
\multirow{7}{*}{\parbox{2cm}{\centering Asymmetric mixup}} & \multicolumn{2}{c|}{m = 0.1} & 52.0 & 48.8 & 68.7 & 32.2 \\
& \multicolumn{2}{c|}{m = 0.15} & 58.0 & 55.7 & 70.6 & 31.6 \\
& \multicolumn{2}{c|}{m = 0.2} & 59.3 & 57.9 & 70.6 & 27.8 \\
& \multicolumn{2}{c|}{m = 0.25} & 60.1 & 55.9 & 78.0 & 23.5 \\
& \multicolumn{2}{c|}{m = 0.3} & 59.2 & 55.4 & 77.9 & 17.6 \\
& \multicolumn{2}{c|}{m = 0.4} & 52.8 & 45.3 & 80.2 & 21.5 \\
& \multicolumn{2}{c|}{m = 0.8} & 51.0 & 43.6 & 79.2 & 18.7 \\
\hline
\end{tabular}
\end{lrbox}
\scalebox{1}{\usebox{\tableboxsens}}
\end{table}

\section{The intensity histogram of different datasets}

\begin{figure}[ht]
\centering
\includegraphics[width=0.44\textwidth]{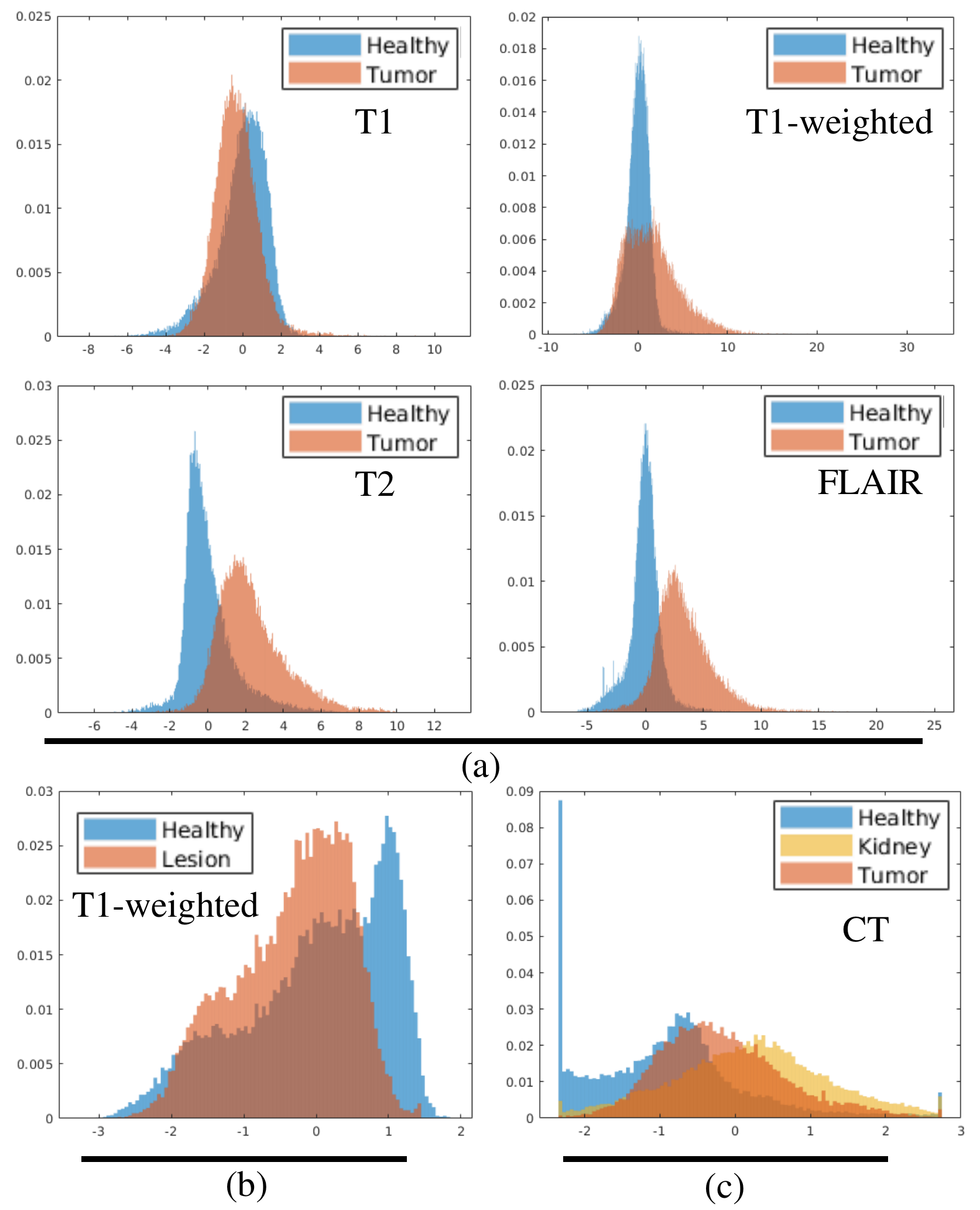}
\caption{(a) The intensity histogram of BRATS, (b) ATLAS and (c) KiTS. The intensity of the foreground and background classes overlap a lot for ATLAS and KiTS. This can be a potential factor due to which the asymmetric mixup does not create useful synthetic samples and cannot improve the segmentation performance that much.} \label{figa7}
\end{figure}

Empirically, we find the asymmetric mixup is the most effective method for tumor segmentation with BRATS. However, asymmetric mixup show limited improvements for ATLAS and KiTS. We think it is because the multi-channel information in BRATS could create more useful information, as shown in Figure \ref{figa7}.

\section{The quantitative results of abdominal organ segmentation}

\begin{figure}[ht]
\centering
\includegraphics[width=0.44\textwidth]{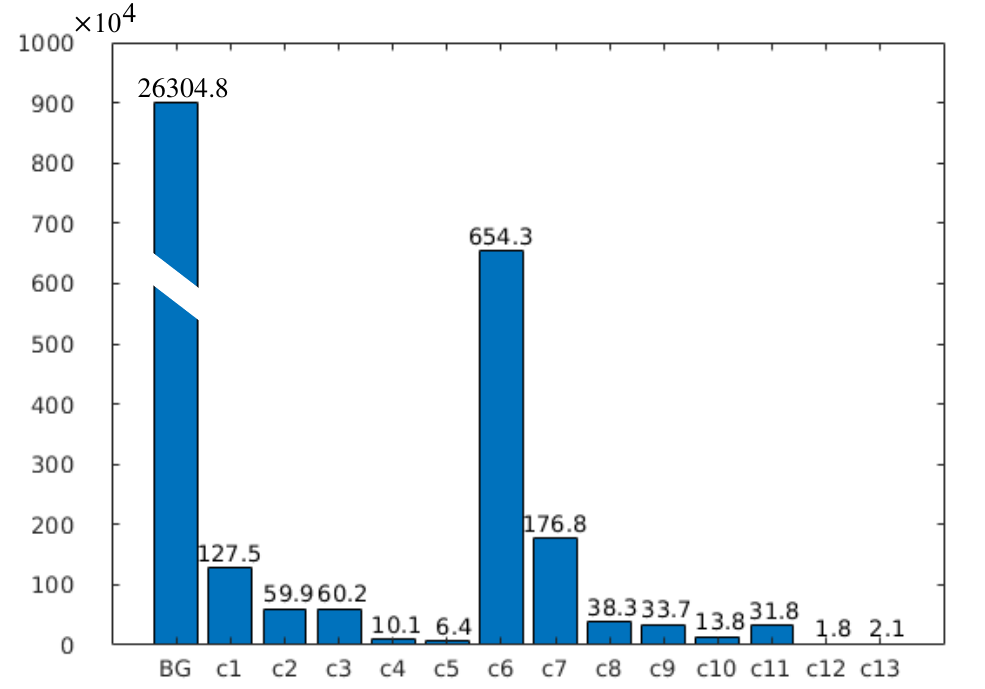}
\caption{The class distribution of the abdomen dataset we use in this study. We summarize the total pixel number of different classes. We take class 4, 5, 8, 9, 10, 11, 12 and 13 as rare classes.} \label{figa8}
\end{figure}

We evaluate one of our proposed techniques, asymmetric focal loss, with the application of abdominal organ segmentation to demonstrate our method can be feasibly applied to multi-class segmentation. Specifically, we train a model of basic DeepMedic using 25\% of the training data, with the same setting in empirical experiments in Section~\ref{sec:motivation}. Considering the class distribution of the dataset, as shown in Figure~\ref{figa8}, we take class 4, class 5, class 8, class 9, class 10, class 11, class 12 and class 13 as rare classes. Specifically, we initiate the one-hot vector $\boldsymbol{r}$ as ${\left[0,0,0,0,1,1,0,0,1,1,1,1,1,1\right]}^\intercal$. We use $\gamma = 4$ in this experiments. We adopt post-processing described in Section~\ref{sec4} separately to the results of every classes. The results are shown in Table~\ref{taba8}. The asymmetric focal loss can get better overall segmentation results than cross entropy or its symmetric variant. More importantly. it can get better segmentation results with higher sensitivity for most rare classes. Specifically, asymmetric focal loss can improve the average DSC of rare classes by 4.9\%. We also notice that asymmetric focal loss would decrease the segmentation performance of esophagus which is taken as a rare class. It is because esophagus is too small, and post-processing would remove the correct segmentation regions by mistake but leave the false positive predictions. We think more advanced post-processing would help improve the segmentation in this case.

\begin{table*}
\centering
\caption{Evaluation of abdomen segmentation with 25\% of training data with symmetric (sy.) and asymmetric (asy.) focal loss. The rare classes are marked with $\boldsymbol{r}$. AVG is the average performance of all classes. AVG$_{\boldsymbol{r}}$ is the average performance of all rare classes. Best results are highlighted in bold.}
\label{taba8}
\newsavebox{\tableboxorgan}
\begin{lrbox}{\tableboxorgan}
\begin{tabular}{p{13mm}<{\centering}|p{7mm}<{\centering}|p{7mm}<{\centering}|p{7mm}<{\centering}|p{7mm}<{\centering}|p{7mm}<{\centering}|p{7mm}<{\centering}|p{7mm}<{\centering}|p{7mm}<{\centering}|p{7mm}<{\centering}|p{7mm}<{\centering}|p{7mm}<{\centering}|p{7mm}<{\centering}|p{7mm}<{\centering}|p{7mm}<{\centering}|p{7mm}<{\centering}}
\hlineB{3}
&  \multicolumn{3}{c|}{c1 (spleen)} & \multicolumn{3}{c|}{c2 (right kidney)} & \multicolumn{3}{c|}{c3 (left kidney)} & \multicolumn{3}{c|}{c4 (gallbladder) $\boldsymbol{r}$} & \multicolumn{3}{c}{c5 (esophagus) $\boldsymbol{r}$} \\
& \multicolumn{1}{m{7mm}<{\centering}|}{vanilla - CE} & \multicolumn{1}{m{7mm}<{\centering}|}{sy. focal loss} & \multicolumn{1}{m{7mm}<{\centering}|}{asy. focal loss} &  \multicolumn{1}{m{7mm}<{\centering}|}{vanilla - CE} & \multicolumn{1}{m{7mm}<{\centering}|}{sy. focal loss} & \multicolumn{1}{m{7mm}<{\centering}|}{asy. focal loss} & \multicolumn{1}{m{7mm}<{\centering}|}{vanilla - CE} & \multicolumn{1}{m{7mm}<{\centering}|}{sy. focal loss} & \multicolumn{1}{m{7mm}<{\centering}|}{asy. focal loss} & \multicolumn{1}{m{7mm}<{\centering}|}{vanilla - CE} & \multicolumn{1}{m{7mm}<{\centering}|}{sy. focal loss} & \multicolumn{1}{m{7mm}<{\centering}|}{asy. focal loss} & \multicolumn{1}{m{7mm}<{\centering}|}{vanilla - CE} & \multicolumn{1}{m{7mm}<{\centering}|}{sy. focal loss} & \multicolumn{1}{m{7mm}<{\centering}}{asy. focal loss} \\
\hlineB{2}
DSC &  \textbf{85.3} & 78.7 & 84.3 & 78.3 & \textbf{84.4} & 68.4 & 82.7 & \textbf{85.0} & 82.6 & 34.6 & 30.5 & \textbf{53.7} & \textbf{55.7} & 52.7 & 41.8  \\
Sensitivity & 80.0 & 70.5 & 76.9 & 73.7 & 77.2 & 62.9 & 77.5 & 79.5 & 77.8 & 25.8 & 22.9 & 46.3 & 50.7 & 48.2 & 43.8 \\
Precision   &  93.6 & 96.0 & 95.9 & 84.6 & 94.8 & 76.2 & 94.7 & 94.4 & 93.0 & 78.4 & 68.5 & 68.9 & 74.4 & 69.5 & 45.0  \\
\hlineB{3}
&  \multicolumn{3}{c|}{c6 (liver)} & \multicolumn{3}{c|}{c7 (stomach)} & \multicolumn{3}{c|}{c8 (aorta) $\boldsymbol{r}$} & \multicolumn{3}{c|}{c9 (vena cava) $\boldsymbol{r}$} & \multicolumn{3}{c}{c10 (vein) $\boldsymbol{r}$} \\
& \multicolumn{1}{m{7mm}<{\centering}|}{vanilla - CE} & \multicolumn{1}{m{7mm}<{\centering}|}{sy. focal loss} & \multicolumn{1}{m{7mm}<{\centering}|}{asy. focal loss} & \multicolumn{1}{m{7mm}<{\centering}|}{vanilla - CE} & \multicolumn{1}{m{7mm}<{\centering}|}{sy. focal loss} & \multicolumn{1}{m{7mm}<{\centering}|}{asy. focal loss} & \multicolumn{1}{m{7mm}<{\centering}|}{vanilla - CE} & \multicolumn{1}{m{7mm}<{\centering}|}{sy. focal loss} & \multicolumn{1}{m{7mm}<{\centering}|}{asy. focal loss} & \multicolumn{1}{m{7mm}<{\centering}|}{vanilla - CE} & \multicolumn{1}{m{7mm}<{\centering}|}{sy. focal loss} & \multicolumn{1}{m{7mm}<{\centering}|}{asy. focal loss} & \multicolumn{1}{m{7mm}<{\centering}|}{vanilla - CE} & \multicolumn{1}{m{7mm}<{\centering}|}{sy. focal loss} & \multicolumn{1}{m{7mm}<{\centering}}{asy. focal loss} \\
\hlineB{2}
DSC          & 87.4 & 88.4 & \textbf{88.6} & 39.7 & 39.7 & \textbf{42.1} & 82.7 & 80.3 & \textbf{84.0} & 65.1 & 66.4 & \textbf{73.9} & 42.0 & 26.3 & \textbf{43.3} \\
Sensitivity  & 84.1 & 85.2 & 84.0 & 28.8 & 28.5 & 31.0 & 76.6 & 73.1 & 82.9 & 56.8 & 60.3 & 76.9 & 28.4 & 16.5 & 31.0 \\
Precision    & 92.3 & 92.8 & 94.2 & 91.3 & 84.1 & 86.3 & 91.6 & 91.5 &  86.3 & 86.1 & 79.3 & 72.7 & 91.5 & 82.1 & 80.0  \\
\hlineB{3}
&  \multicolumn{3}{c|}{c11 (pancreas) $\boldsymbol{r}$} & \multicolumn{3}{c|}{c12 (right adrenal) $\boldsymbol{r}$} & \multicolumn{3}{c|}{c13 (left adrenal) $\boldsymbol{r}$} & \multicolumn{3}{c|}{AVG} & \multicolumn{2}{c}{AVG$_{\boldsymbol{r}}$}  \\
& \multicolumn{1}{m{7mm}<{\centering}|}{vanilla - CE} & \multicolumn{1}{m{7mm}<{\centering}|}{sy. focal loss} & \multicolumn{1}{m{7mm}<{\centering}|}{asy. focal loss} & \multicolumn{1}{m{7mm}<{\centering}|}{vanilla - CE} & \multicolumn{1}{m{7mm}<{\centering}|}{sy. focal loss} & \multicolumn{1}{m{7mm}<{\centering}|}{asy. focal loss} & \multicolumn{1}{m{7mm}<{\centering}|}{vanilla - CE} & \multicolumn{1}{m{7mm}<{\centering}|}{sy. focal loss} & \multicolumn{1}{m{7mm}<{\centering}|}{asy. focal loss} & \multicolumn{1}{m{7mm}<{\centering}|}{vanilla - CE} & \multicolumn{1}{m{7mm}<{\centering}|}{sy. focal loss} & \multicolumn{1}{m{7mm}<{\centering}|}{asy. focal loss} & \multicolumn{1}{m{7mm}<{\centering}|}{vanilla - CE} & \multicolumn{1}{m{7mm}<{\centering}|}{sy. focal loss} & \multicolumn{1}{m{7mm}<{\centering}}{asy. focal loss} \\
\hlineB{2}
DSC & 17.2 & 24.3 & \textbf{26.4} & 54.3 & 32.8 & \textbf{55.9} & 34.8 & 28.9 & \textbf{47.0} & 58.5 & 55.3 & \textbf{60.9} & 48.3 & 42.8 & \textbf{53.2} \\
Sensitivity & 11.1 & 17.3 & 18.3 & 45.3 & 24.3 & 50.3 & 27.2 & 22.2 & 41.6 & 51.2 & 48.1 & 55.7 & 40.2 & 35.6 & 48.9 \\
Precision   & 56.6 & 52.0 & 61.7 & 74.5 & 60.8 & 69.4 & 61.4 & 52.7 & 63.6 & 82.4 & 78.3 & 76.4 & 76.8 & 69.6 & 68.5 \\
\hline
\end{tabular}
\end{lrbox}
\scalebox{0.95}{\usebox{\tableboxorgan}}
\end{table*}

\section{Quantitative results without post-processing}

\begin{table*}[t]
\centering
\caption{Evaluation of brain tumor core segmentation using DeepMedic with different amounts of training data and different techniques to counter overfitting. The results are calculated without any post-processing. Results which have worse DSC than the vanilla baseline are highlighted with gray shading. Best and second best results are highlighted in bold with the best also being underlined.}\label{taba4}
\newsavebox{\tableboxbt}
\begin{lrbox}{\tableboxbt}
\begin{tabular}{p{38mm}<{\centering}|c|c|c|c||c|c|c|c||c|c|c|c||c|c|c|c}
\hlineB{3}
\multirow{2}{*}{Method} & \multicolumn{4}{c||}{5\% training} & \multicolumn{4}{c||}{10\% training} & \multicolumn{4}{c||}{20\% training} & \multicolumn{4}{c}{50\% training}  \\ 
 &  DSC & SEN &  PRC & HD &   DSC & SEN &  PRC & HD &   DSC & SEN &  PRC & HD &   DSC & SEN &  PRC & HD  \\
\hlineB{2}
Vanilla - CE~\cite{kamnitsas2017efficient} &  51.0  & 42.6 &  78.6 & 17.5 &  62.8  & 56.9 &  81.6 & \textbf{\underline{13.7}} &  65.3  & 61.0 & 83.2 & 12.8 & 69.5 & 66.4 & 83.8 & 14.3 \\
Vanilla - CE - 80\% tumor &  \cellcolor{gray!25}46.2  & \cellcolor{gray!25}38.3 &  \cellcolor{gray!25}77.1 & \cellcolor{gray!25}22.5 &  \cellcolor{gray!25}61.6  & \cellcolor{gray!25}55.3 &  \cellcolor{gray!25}79.1 & \cellcolor{gray!25}17.8 & 65.5  & 60.9 & 81.7 & 17.1 & \cellcolor{gray!25}68.8 & \cellcolor{gray!25}65.3 & \cellcolor{gray!25}83.4 & \cellcolor{gray!25}15.1 \\
Vanilla - F1 (DSC) &  \cellcolor{gray!25}47.7  & \cellcolor{gray!25}38.7 &  \cellcolor{gray!25}82.0 & \cellcolor{gray!25}\textbf{15.4} &  \cellcolor{gray!25}59.4  & \cellcolor{gray!25}52.2 & \cellcolor{gray!25}82.7 & \cellcolor{gray!25}\textbf{\underline{13.7}} &  \cellcolor{gray!25}64.6  & \cellcolor{gray!25}59.0 & \cellcolor{gray!25}82.9 & \cellcolor{gray!25}13.6 & \cellcolor{gray!25}67.4 & \cellcolor{gray!25}63.4 & \cellcolor{gray!25}84.7 & \cellcolor{gray!25}13.4 \\
Vanilla - F2~\cite{hashemi2018asymmetric} & \cellcolor{gray!25}46.9 & \cellcolor{gray!25}38.6 & \cellcolor{gray!25}79.2 & \cellcolor{gray!25}\textbf{\underline{14.9}} & \cellcolor{gray!25}59.9 & \cellcolor{gray!25}53.6 & \cellcolor{gray!25}82.6 & \cellcolor{gray!25}15.4 & 66.5 & 62.1 & 81.8 & \textbf{12.7} & \cellcolor{gray!25}68.8 & \cellcolor{gray!25}67.0 & \cellcolor{gray!25}81.1 & \cellcolor{gray!25}14.2 \\
Vanilla - F4~\cite{hashemi2018asymmetric} & 51.6 & 44.0 & 78.6 & 19.8 & \cellcolor{gray!25}60.1 & \cellcolor{gray!25}54.1 & \cellcolor{gray!25}81.7 & \cellcolor{gray!25}16.5 & 65.9 & 63.1 & 80.9 & 19.1 & \cellcolor{gray!25}67.8 & \cellcolor{gray!25}65.7 & \cellcolor{gray!25}83.1 & \cellcolor{gray!25}12.2 \\
Vanilla - F8~\cite{hashemi2018asymmetric} & \cellcolor{gray!25}48.6 & \cellcolor{gray!25}40.2 & \cellcolor{gray!25}80.2 & \cellcolor{gray!25}16.8 & \cellcolor{gray!25}60.2 & \cellcolor{gray!25}53.6 & \cellcolor{gray!25}83.2 & \cellcolor{gray!25}15.4 & \cellcolor{gray!25}64.7 & \cellcolor{gray!25}61.3 & \cellcolor{gray!25}81.6 & \cellcolor{gray!25}15.4 & \cellcolor{gray!25}67.9 & \cellcolor{gray!25}66.4 & \cellcolor{gray!25}79.6 & \cellcolor{gray!25}13.7 \\
\hline
Large margin loss~\cite{liu2016large} &  \cellcolor{gray!25}46.4 & \cellcolor{gray!25}38.3 & \cellcolor{gray!25}77.8 & \cellcolor{gray!25}21.3 & \cellcolor{gray!25}61.2 & \cellcolor{gray!25}54.3 & \cellcolor{gray!25}82.2 & \cellcolor{gray!25}15.3 & 67.0  & 62.7 &  83.3 & \textbf{\underline{12.5}} & \cellcolor{gray!25}66.8 & \cellcolor{gray!25}63.4 & \cellcolor{gray!25}86.1 & \cellcolor{gray!25}\textbf{\underline{11.0}} \\
Asymmetric large margin loss &  55.5 & 50.6 & 76.2 & 23.9 & 64.3  & 57.9 & 84.1&  \textbf{14.3} &  67.8  & 63.9 & 82.4 & 13.2 &  \cellcolor{gray!25}69.3 & \cellcolor{gray!25}66.2 & \cellcolor{gray!25}84.6 & \cellcolor{gray!25}12.7  \\
\hline
Focal loss~\cite{lin2017focal} & 53.6  & 46.3 & 78.7 & 20.3 &  62.9  & 56.0 &  82.5 & 17.3 &  \cellcolor{gray!25}65.2  & \cellcolor{gray!25}61.1 & \cellcolor{gray!25}82.6 & \cellcolor{gray!25}19.2 & \cellcolor{gray!25}67.2 & \cellcolor{gray!25}63.2 & \cellcolor{gray!25}84.7& \cellcolor{gray!25}15.3 \\
Asymmetric focal loss & 57.3 & 52.7 &  74.4& 24.4 & 66.3   & 62.9 & 79.1 & 16.5 & 68.6  & 67.3 & 78.8 & 15.6 & \textbf{71.2} & 71.7 & 79.9 & 12.4 \\
\hline
Adversarial training~\cite{goodfellow2014explaining} &  53.4  & 45.7 &  81.8 & 21.5 &  \cellcolor{gray!25}62.4 & \cellcolor{gray!25}55.8 &  \cellcolor{gray!25}83.1 &  \cellcolor{gray!25}19.4 &  \cellcolor{gray!25}65.2  & \cellcolor{gray!25}60.4 & \cellcolor{gray!25}83.4 & \cellcolor{gray!25}15.4 & \cellcolor{gray!25}66.0 & \cellcolor{gray!25}62.0 & \cellcolor{gray!25}84.8 & \cellcolor{gray!25}17.8 \\
Asymmetric adversarial training & 56.8 & 51.8 & 74.8 & 23.6 & 64.0  & 59.2 & 80.5 & 17.2 & 68.0  & 64.7 & 82.8 & 15.6 & 70.6 & 69.3 & 81.5 & 15.0\\
\hline
Mixup~\cite{zhang2017mixup} & \cellcolor{gray!25}50.0 & \cellcolor{gray!25}42.2 & \cellcolor{gray!25}77.6 & \cellcolor{gray!25}21.1  & \cellcolor{gray!25}60.9 & \cellcolor{gray!25}55.0 & \cellcolor{gray!25}81.4 & \cellcolor{gray!25}19.7 & \cellcolor{gray!25}64.9 & \cellcolor{gray!25}60.0 &  \cellcolor{gray!25}82.3 & \cellcolor{gray!25}17.3 &
\cellcolor{gray!25}67.2 & \cellcolor{gray!25}62.7 & \cellcolor{gray!25}86.3 & \cellcolor{gray!25}17.3 \\
Asymmetric mixup &  \textbf{59.2} &  57.9  & 70.7 & 27.8  &  \textbf{68.5}  & 66.3 & 79.2 & 16.5 &  \textbf{70.6}  & 69.2 & 81.2 & 16.0 & 70.8 & 69.1 & 83.7 & \textbf{11.1} \\
\hline
Symmetric combination & \cellcolor{gray!25}50.6 & \cellcolor{gray!25}43.0 & \cellcolor{gray!25}82.2 & \cellcolor{gray!25}20.3 & \cellcolor{gray!25}61.0 & \cellcolor{gray!25}54.2 & \cellcolor{gray!25}83.4 & \cellcolor{gray!25}23.3 & \cellcolor{gray!25}64.9 & \cellcolor{gray!25}59.5 & \cellcolor{gray!25}85.9 & \cellcolor{gray!25}16.8 & \cellcolor{gray!25}67.4 & \cellcolor{gray!25}63.9 & \cellcolor{gray!25}84.4 & \cellcolor{gray!25}15.7 \\
Asymmetric combination &  \textbf{\underline{62.4}}  & 64.7 &  71.5 & 27.8  &  \textbf{\underline{71.4}} & 73.8 &  74.3 & 20.8 &  \textbf{\underline{71.9}}  & 74.1 &  79.2 & 20.7 & \textbf{\underline{72.9}} & 77.0 & 77.8 & 19.0 \\
\hline
\end{tabular}
\end{lrbox}
\scalebox{0.91}{\usebox{\tableboxbt}}
\end{table*}

\begin{table*}[t]
\centering
\caption{Evaluation of brain stroke lesion segmentation on ATLAS using DeepMedic with different amounts of training data and different techniques to counter overfitting. The results are calculated without post-processing. Results which have worse DSC than the vanilla baseline are highlighted with shading. Best and second best results are in bold with the best also underlined.}\label{taba5}
\newsavebox{\tableboxatlassl}
\begin{lrbox}{\tableboxatlassl}
\begin{tabular}{p{43mm}<{\centering}|c|c|c|c||c|c|c|c||c|c|c|c}
\hlineB{3}
\multirow{2}{*}{Method} & \multicolumn{4}{c||}{30\% training} & \multicolumn{4}{c||}{50\% training} & \multicolumn{4}{c}{100\% training}  \\ 
 &  DSC & SEN &  PRC & HD &   DSC & SEN &  PRC & HD & DSC & SEN &  PRC & HD \\
\hlineB{2}
Vanilla - w/ augmentation~\cite{kamnitsas2017efficient} &  22.9 & 22.4 & 52.3 & 41.7 & 47.7 & 48.5 & 55.3 & \textbf{32.8}  & 55.7 & 56.8 & 62.2 & 30.2 \\
Vanilla - w/o augmentation & 17.1 & 15.2 & 51.4 & 38.8 & 40.5 & 46.8 & 45.8 & 46.4 & 53.5 & 56.0 & 58.2 & 33.3 \\
Vanilla - asymmetric augmentation & 23.3 & 22.6 & 49.7 & 41.8 & 48.7 & 51.3 & 55.0 & 35.0 & \textbf{57.1} & 58.9 & 62.7 & \textbf{29.1} \\
\hline
Large margin loss~\cite{liu2016large} & \cellcolor{gray!25}20.2 & \cellcolor{gray!25}17.0 & \cellcolor{gray!25}59.4 & \cellcolor{gray!25}\textbf{\underline{37.2}} & \cellcolor{gray!25}46.8 & \cellcolor{gray!25}46.2 & \cellcolor{gray!25}57.4 & \cellcolor{gray!25}34.9 & 56.0 & 55.3 & 64.3 & \textbf{\underline{29.0}} \\
Asymmetric large margin loss & 24.0 & 23.8 & 50.4 & 41.4 & 49.2 & 52.6 & 54.7 & 35.4 & 56.9 & 59.9 & 60.8 & 27.7 \\
\hline
Focal loss~\cite{lin2017focal} & \cellcolor{gray!25}21.9 & \cellcolor{gray!25}20.2 & \cellcolor{gray!25}55.0 & \cellcolor{gray!25}\textbf{37.7} & 47.8 & 49.3 & 55.6 & 33.7 & \textbf{57.1} & 59.6 & 63.4 & 32.6 \\
Asymmetric focal loss & 24.7 & 27.2 & 42.8 & 49.0 & 49.6 & 56.3 & 51.5 & 36.6 & 56.6 & 64.7 & 56.9 & 31.0 \\
\hline
Adversarial training~\cite{goodfellow2014explaining} & \cellcolor{gray!25}21.1 & \cellcolor{gray!25}18.3 & \cellcolor{gray!25}55.1& \cellcolor{gray!25}49.2 & 48.3 & 46.1 & 59.1 & 35.9 & 56.4 & 55.2 & 65.2 & 34.0 \\
Asymmetric adversarial training & \textbf{27.5} & 28.1 & 52.3 & 42.2 & \textbf{50.8} & 52.6 & 57.6 & 33.3 & 56.7 & 58.5 & 64.3 & 33.4 \\
\hline
Mixup~\cite{zhang2017mixup} & \cellcolor{gray!25}15.9 & \cellcolor{gray!25}14.9 & \cellcolor{gray!25}46.0 & \cellcolor{gray!25}41.0 & \cellcolor{gray!25}47.6 & \cellcolor{gray!25}47.5 & \cellcolor{gray!25}56.8 & \cellcolor{gray!25}\textbf{\underline{31.6}} & 55.9 & 57.4 & 63.6 & 30.3 \\
Asymmetric mixup & \cellcolor{gray!25}21.5 & \cellcolor{gray!25}24.9 & \cellcolor{gray!25}39.4 & \cellcolor{gray!25}48.0 & 47.8 & 60.3 & 46.3 & 45.8 & 57.0 & 56.3 & 67.1 & 33.8 \\
\hline
Symmetric combination & 24.6 & 20.9 & 63.5 & 41.3 & 49.7 & 51.7 & 56.0 & 34.7 & 56.8 & 57.0 & 65.1 & 29.9 \\
Asymmetric combination & \textbf{\underline{29.9}} & 34.2 & 47.1 & 47.4 & \textbf{\underline{51.1}} & 58.6 & 52.2 & 38.6 & \textbf{\underline{57.9}} & 62.4 & 61.5 & 32.1 \\

\hline
\end{tabular}
\end{lrbox}
\scalebox{1}{\usebox{\tableboxatlassl}}
\end{table*}

\begin{table*}[t]
\centering
\caption{Evaluation of kidney and kidney tumor segmentation based on 3D U-Net with different amounts of training data and different techniques to counter overfitting. The results are calculated without any post-processing. Results which have worse DSC than the vanilla baseline are highlighted with shading. Best and second best results are in bold with the best also underlined.}\label{taba6}
\newsavebox{\tableboxkitss}
\begin{lrbox}{\tableboxkitss}
\begin{tabular}{p{40mm}<{\centering}|p{4mm}<{\centering}|p{4mm}<{\centering}|p{4mm}<{\centering}|p{4mm}<{\centering}||p{4mm}<{\centering}|p{4mm}<{\centering}|p{4mm}<{\centering}|p{4mm}<{\centering}||p{4mm}<{\centering}|p{4mm}<{\centering}|p{4mm}<{\centering}|p{4mm}<{\centering}}
\hlineB{3}
\multirow{3}{*}{Method} & \multicolumn{12}{c}{Kidney} \\
 & \multicolumn{4}{c||}{10\% training} & \multicolumn{4}{c||}{50\% training} & \multicolumn{4}{c}{100\% training} \\
 & DSC & SEN &  PRC & HD &  DSC & SEN &  PRC & HD &  DSC & SEN &  PRC & HD \\
\hlineB{2}
Vanilla - w/ augmentation~\cite{isensee2019automated} & 93.7 & 91.7 & 96.9 & 5.6 & 96.5 & 96.1 & 97.0 & 3.6 & \textbf{96.8} & 96.4 & 97.2 & \textbf{2.2}  \\
Vanilla - w/o augmentation &  92.8 & 90.2 & 96.6 & 12.4 & 96.3 & 93.1 & 96.6 & 2.5 & 96.5 & 96.4 & 96.8 & 3.8 \\
Vanilla - asymmetric augmentation & 94.7 & 93.0 & 96.9 & 5.2 & 95.6 & 95.9 & 95.8 & 5.9 & 96.5 & 96.6 & 96.6 & 3.7 \\
\hline
Large margin loss~\cite{liu2016large} & \textbf{94.9} & 93.1 & 97.0 & 4.7 & 96.4 & 96.3 & 96.7 & 4.0 & 96.3 & 96.7 & 96.1 & 4.6 \\
Asymmetric large margin loss &  94.1 & 91.9 & 97.1 & 5.9 & 96.3 & 95.9 & 96.8 & 3.2 & \textbf{96.8} & 96.8 & 96.8 & 4.0 \\
\hline
Focal loss~\cite{lin2017focal} & 91.6 & 86.1 & 99.1 & 6.6 & 94.1 & 89.9 & 99.0 & 5.5 & 94.4 & 90.2 & 99.1 & 4.1 \\
Asymmetric focal loss & 92.4 & 87.3 & 98.9 & 5.8 & 94.9 & 91.3 & 98.9 & 3.4 & 94.9 & 91.2 & 99.1 & 3.0 \\
\hline
Adversarial training~\cite{goodfellow2014explaining} &  94.3 & 92.3 & 97.3 & 6.3 & 96.5 & 96.1 & 97.0 & \textbf{2.4} & \textbf{96.8} & 96.5 & 97.2 & \textbf{2.2} \\ 
Asymmetric adversarial training & 94.6 & 92.8 & 97.2 & 4.6 & \textbf{96.6} & 93.4 & 97.0 & 3.6 & \textbf{\underline{97.0}} & 96.7 & 97.3 & \textbf{\underline{2.1}} \\
\hline
Mixup~\cite{zhang2017mixup} & \textbf{\underline{95.2}} & 93.6 & 97.3 & \textbf{\underline{4.0}} & \textbf{\underline{96.9}} & 96.4 & 97.5 & \textbf{\underline{2.2}} & \textbf{\underline{97.0}} & 96.6 & 97.3 & 2.5 \\
Asymmetric mixup & 94.8 & 92.9 & 97.3 & \textbf{4.3} & 96.1 & 95.4 & 97.0 & 3.1 & 96.5 & 95.9 & 97.3 & 2.5 \\
\hline
Symmetric combination & 94.3 & 91.9 & 97.4 & 5.9 & 94.7 & 91.2 & 98.7 & 4.1 & \textbf{96.8} & 96.4 & 97.2 & \textbf{\underline{2.1}} \\
Asymmetric combination & 94.0 & 90.5 & 98.4 & 4.8 & 94.3 & 90.8 & 95.4 & 5.1 & \textbf{96.8} & 95.9 & 97.7 & 3.3 \\
\hlineB{3}
\multirow{3}{*}{Method} & \multicolumn{12}{c}{Kidney tumor} \\
 & \multicolumn{4}{c||}{10\% training} & \multicolumn{4}{c||}{50\% training} & \multicolumn{4}{c}{100\% training} \\
 & DSC & SEN &  PRC & HD &  DSC & SEN &  PRC & HD &  DSC & SEN &  PRC & HD \\
\hlineB{2}
Vanilla - w/ augmentation~\cite{isensee2019automated} & 54.9 & 47.9 & 77.5 & 93.8 & 75.6 & 73.7 & 83.9 & 49.8 & 79.3 & 78.3 & 84.8 & 48.3  \\
Vanilla - w/o augmentation &  37.8 & 32.9 & 62.8 & 121.2 & 64.1 & 62.1 & 74.6 & 85.6 & 71.3 & 69.7 & 79.3 & 54.9 \\
Vanilla - asymmetric augmentation & 56.1 & 50.4 & 74.5 & 97.1 & \cellcolor{gray!25}75.3 & \cellcolor{gray!25}73.6 & \cellcolor{gray!25}84.3 & \cellcolor{gray!25}60.3 & 79.6 & 79.5 & 85.2 & \textbf{35.0} \\
\hline
Large margin loss~\cite{liu2016large} & \cellcolor{gray!25}54.8 & \cellcolor{gray!25}48.2 & \cellcolor{gray!25}77.2 & \cellcolor{gray!25}84.0 & 77.1 & 75.2 & 84.5 & 58.8 & 80.9 & 82.1 & 83.5 & 47.4 \\
Asymmetric large margin loss & 55.7 & 50.1 & 75.4 & 99.6 & 77.9 & 76.0 & 84.9 & 54.6 & \textbf{81.9} & 82.3 & 84.0 & 56.2 \\
\hline
Focal loss~\cite{lin2017focal} & \cellcolor{gray!25}48.4 & \cellcolor{gray!25}39.3 & \cellcolor{gray!25}78.1 & \cellcolor{gray!25}80.7 & \cellcolor{gray!25}73.0 & \cellcolor{gray!25}66.8 & \cellcolor{gray!25}86.1 & \cellcolor{gray!25}63.0 & \cellcolor{gray!25}78.6 & \cellcolor{gray!25}73.6 & \cellcolor{gray!25}88.1 & \cellcolor{gray!25}52.5 \\
Asymmetric focal loss &  \textbf{57.0} & 49.9 & 74.9 & 95.9 & \textbf{78.2} & 76.6 & 84.6 & \textbf{43.8} & 80.8 & 81.1 & 83.5 & 48.9 \\
\hline
Adversarial training~\cite{goodfellow2014explaining} &  \cellcolor{gray!25}51.6 & \cellcolor{gray!25}45.0 & \cellcolor{gray!25}79.5 & \cellcolor{gray!25}\textbf{78.3} & \cellcolor{gray!25}73.6 & \cellcolor{gray!25}71.6 & \cellcolor{gray!25}83.2 & \cellcolor{gray!25}55.3 & 81.4 & 81.6 & 84.0 & 52.2 \\ 
Asymmetric adversarial training & 56.9 & 51.1 & 79.4 & 87.5 & 77.4 & 75.6 & 85.5 & 58.3 & 81.8 & 81.5 & 86.1 & \textbf{\underline{30.8}} \\
\hline
Mixup~\cite{zhang2017mixup} & \cellcolor{gray!25}54.5 & \cellcolor{gray!25}48.3 & \cellcolor{gray!25}79.6 & \cellcolor{gray!25}\textbf{\underline{74.3}} & 77.1 & 73.8 & 86.2 & 48.3 & 80.6 & 79.5 & 84.9 & 52.0 \\
Asymmetric mixup & 55.1& 48.6 & 79.9 & 92.3 & 77.9 & 74.4 & 87.9 & \textbf{\underline{40.8}} & 79.8 & 79.0 & 85.9  & 54.9 \\
\hline
Symmetric combination & \cellcolor{gray!25}54.2 & \cellcolor{gray!25}47.1 & \cellcolor{gray!25}79.0 & \cellcolor{gray!25}105.4 & \cellcolor{gray!25}73.6 & \cellcolor{gray!25}67.5 & \cellcolor{gray!25}86.0 & \cellcolor{gray!25}51.5 & 80.5 & 80.0 & 84.5 & 48.4 \\
Asymmetric combination & \textbf{\underline{59.2}} & 54.1 & 77.1 & 82.8 & \textbf{\underline{79.4}} & 79.0 & 85.1 & 50.6 & \textbf{\underline{82.2}} & 82.7 & 85.0 & 36.7 \\
\hline
\end{tabular}
\end{lrbox}
\scalebox{1}{\usebox{\tableboxkitss}}
\end{table*}

The quantitative segmentation results without post-processing are summarized in Table~\ref{taba4}, Table~\ref{taba5} and Table~\ref{taba6}. Without post-processing, the proposed asymmetric regularization methods can improve DSC but could lead to worse distance-based evaluation metrics such as Hausdorff distance (HD). It is because the regularized model, which is more sensitive for the under-represented classes, would make relatively more false positive predictions. The false positive predictions which are far from the ground truth would increase HD significantly. However, in practice most false positive predictions could be easily removed by some connected component-based post-processing, as described in Section~\ref{sec4}. In this way, eventually we can get better or similar HD with our methods, as shown in the main text.


\end{document}